\documentclass{article}

\usepackage[utf8]{inputenc} \usepackage[T1]{fontenc}    \usepackage{hyperref}       \usepackage{url}            \usepackage{booktabs}       \usepackage{amsfonts}       \usepackage{nicefrac}       \usepackage{color}
\usepackage{epsfig}
\usepackage{amsthm}
\usepackage{amsmath}
\usepackage{float}
\usepackage{graphicx}
\usepackage{wrapfig}
\usepackage{makecell}
\usepackage{caption}
\usepackage{subcaption}
\usepackage{bm}
\usepackage{multirow}
\usepackage{microtype}
\usepackage{mathtools}
\usepackage{tikz}
\usepackage{gensymb}
\usepackage{bbm}
\usepackage{chngcntr}
\usepackage{amssymb}
\usepackage{enumitem}
\usepackage{placeins}

\usepackage{hyperref}

\usepackage[accepted]{icml2020}

\newcommand{\one}[1]{\mathbbm{1}_{[#1]}}

\begin{document}

\twocolumn[
\icmltitle{A Simple Framework for Contrastive Learning of Visual Representations}

\icmlsetsymbol{equal}{*}

\begin{icmlauthorlist}
\icmlauthor{Ting Chen}{goo}
\icmlauthor{Simon Kornblith}{goo}
\icmlauthor{Mohammad Norouzi}{goo}
\icmlauthor{Geoffrey Hinton}{goo}
\end{icmlauthorlist}

\icmlaffiliation{goo}{Google Research, Brain Team}

\icmlcorrespondingauthor{Ting Chen}{iamtingchen@google.com}

\icmlkeywords{Self-supervised Learning, Contrastive Learning, Deep Learning}

\vskip 0.3in
]
\setlength{\textfloatsep}{0.15in}

\printAffiliationsAndNotice{}

\begin{abstract}
This paper presents \textit{SimCLR}: a simple framework for contrastive learning of visual representations. We simplify recently proposed contrastive self-supervised learning algorithms without requiring specialized architectures or a memory bank. In order to understand what enables the contrastive prediction tasks to learn useful representations, we systematically study the major components of our framework. We show that (1) composition of data augmentations plays a critical role in defining effective predictive tasks, (2) introducing a learnable nonlinear transformation between the representation and the contrastive loss substantially improves the quality of the learned representations, and (3) contrastive learning benefits from larger batch sizes and more training steps compared to supervised learning. By combining these findings, we are able to considerably outperform previous methods for self-supervised and semi-supervised learning on ImageNet. A linear classifier trained on self-supervised representations learned by SimCLR achieves 76.5\% top-1 accuracy, which is a 7\% relative improvement over previous state-of-the-art, matching the performance of a supervised ResNet-50. When fine-tuned on only 1\% of the labels, we achieve 85.8\% top-5 accuracy, outperforming AlexNet with 100$\times$ fewer labels.~\footnote{Code available at \href{https://github.com/google-research/simclr}{https://github.com/google-research/simclr}.}
\end{abstract}
 \section{Introduction}
Learning effective visual representations without human supervision is a long-standing problem.
Most mainstream approaches fall into one of two classes: generative or discriminative. Generative approaches learn to generate or otherwise model pixels in the input space~\cite{hinton2006fast,kingma2013auto,goodfellow2014generative}. However, pixel-level generation is computationally expensive and may not be necessary for representation learning. \linebreak Discriminative approaches learn representations using objective functions similar to those used for supervised learning, but train networks to perform pretext tasks where both the inputs and labels are derived from an unlabeled dataset. Many such approaches have relied on heuristics to design pretext tasks~\cite{doersch2015unsupervised,zhang2016colorful,noroozi2016unsupervised,gidaris2018unsupervised}, which could limit the generality of the learned representations. Discriminative approaches based on contrastive learning in the latent space have recently shown great promise, achieving state-of-the-art results~\cite{hadsell2006dimensionality,dosovitskiy2014discriminative,oord2018representation,bachman2019learning}.

\begin{figure}[t]
    \centering
    \includegraphics[width=\linewidth]{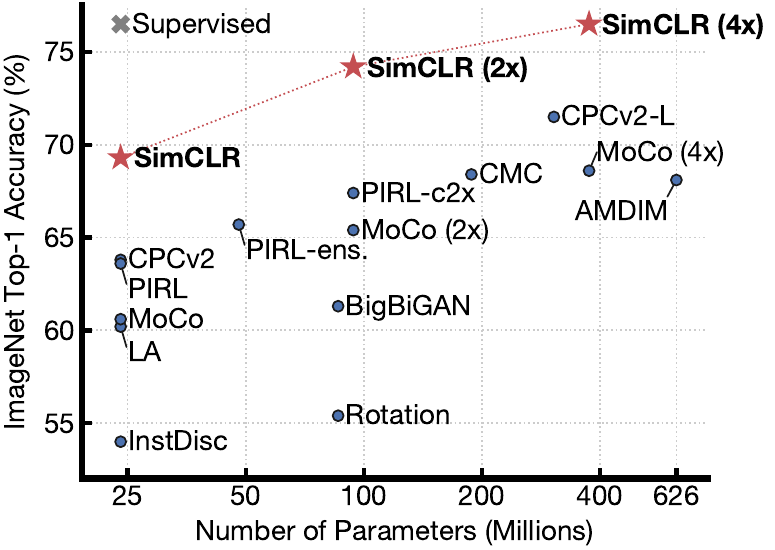}
\caption{\label{fig:linear_sota}
    ImageNet Top-1 accuracy of linear classifiers trained on representations learned with different self-supervised methods (pretrained on ImageNet). Gray cross indicates supervised \mbox{ResNet-50}. Our method, SimCLR, is shown in bold.}
\end{figure}

In this work, we introduce a simple framework for contrastive learning of visual representations, which we call \textit{SimCLR}.
Not only does SimCLR outperform previous work (Figure~\ref{fig:linear_sota}), but it is also simpler, requiring neither specialized architectures~\cite{bachman2019learning,henaff2019data} nor a memory bank~\cite{wu2018unsupervised,tian2019contrastive,he2019momentum,misra2019self}.

In order to understand what enables good contrastive representation learning, we systematically study the major components of our framework and show that:
\begin{itemize}[topsep=0pt, partopsep=0pt, leftmargin=13pt, parsep=0pt, itemsep=4pt]
    \item Composition of multiple data augmentation operations is crucial in defining the contrastive prediction tasks that yield effective representations. In addition, unsupervised contrastive learning benefits from stronger data augmentation than supervised learning.
    \item Introducing a learnable nonlinear transformation between the representation and the contrastive loss substantially improves the quality of the learned representations.
    \item Representation learning with contrastive cross entropy loss benefits from normalized embeddings and an appropriately adjusted temperature parameter.
    \item Contrastive learning benefits from larger batch sizes and longer training compared to its supervised counterpart. Like supervised learning, contrastive learning benefits from deeper and wider networks.
\end{itemize}

We combine these findings to achieve a new state-of-the-art in self-supervised and semi-supervised learning on ImageNet ILSVRC-2012~\cite{russakovsky2015imagenet}. 
Under the linear evaluation protocol, SimCLR achieves 76.5\% top-1 accuracy, which is a 7\% relative improvement over previous state-of-the-art~\cite{henaff2019data}.
When fine-tuned with only 1\% of the ImageNet labels, SimCLR achieves 85.8\% top-5 accuracy, a relative improvement of 10\%~\cite{henaff2019data}.
When fine-tuned on other natural image classification datasets, SimCLR performs on par with or better than a strong supervised baseline~\cite{kornblith2019better} on 10 out of 12 datasets. \section{Method}
\subsection{The Contrastive Learning Framework}

\begin{figure}[!t]
\small
    \centering
\begin{tikzpicture}
    \node at (0,1.8) (h) {$\longleftarrow\,$Representation$\,\longrightarrow$};
    \node[draw, circle] at (0,-1) (x) {$\,~\bm{x}~\,$};
    \node[draw, circle] at (-2.5,0) (x1) {$\tilde{\bm{x}}_i$};
    \node[draw, circle] at (2.5,0) (x2) {$\tilde{\bm{x}}_j$};
    \node at (-2.5,1.8) (h) {$\bm h_i$};
    \node at (2.5,1.8) (c) {$\bm h_j$};
    \node at (-2.5,3) (hh) {$\bm z_i$};
    \node at (2.5,3) (cc) {$\bm z_j$};
    \path[->] 
        (x)  edge [>=latex] node[below,rotate=-25] {$t\sim\mathcal{T}$} (x1)
        (x)  edge [>=latex] node[below,rotate=25] {$t'\sim \mathcal{T}$} (x2)
        (x1)  edge [>=latex] node[left,rotate=0] {$f(\cdot)$} (h)
        (x2)  edge [>=latex] node[right,rotate=0] {$f(\cdot)$} (c)
        (h)  edge [>=latex] node[left,rotate=0] {$g(\cdot)$} (hh)
        (c)  edge [>=latex] node[right,rotate=0] {$g(\cdot)$} (cc);
    \path[<->]
        (hh)  edge [>=latex] node[above,rotate=0] {Maximize agreement} (cc);
    \end{tikzpicture}
    \caption{A simple framework for contrastive learning of visual representations. 
    Two separate data augmentation operators are sampled from the same family of augmentations ($t\sim \mathcal{T}$ and $t'\sim \mathcal{T}$) and applied to each data example to obtain two correlated views.
    A base encoder network $f(\cdot)$ and a projection head $g(\cdot)$ are trained to maximize agreement using a contrastive loss. After training is completed, we throw away the projection head $g(\cdot)$ and use encoder $f(\cdot)$ and representation $\bm h$ for downstream tasks.}
    \label{fig:framework}
\end{figure}
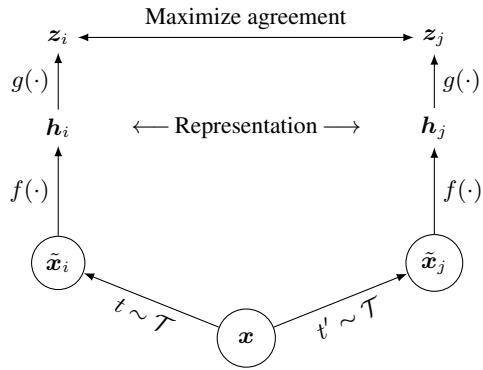

Inspired by recent contrastive learning algorithms (see Section \ref{sec:related} for an overview), SimCLR learns representations by maximizing agreement between differently augmented views of the same data example via a contrastive loss in the latent space. As illustrated in Figure \ref{fig:framework}, this framework comprises the following four major components.
\begin{itemize}[topsep=0pt, partopsep=0pt, leftmargin=13pt, parsep=0pt, itemsep=4pt]
    \item A stochastic \textit{data augmentation} module that transforms any given data example randomly 
    resulting in two correlated views of the same example, denoted $\tilde{\bm x}_i$ and $\tilde{\bm x}_j$, which we consider as a positive pair.
    In this work, we sequentially apply three simple augmentations: \textit{random cropping} followed by resize back to the original size, \textit{random color distortions}, and \textit{random Gaussian blur}. As shown in Section~\ref{sec:da}, the combination of random crop and color distortion is crucial to achieve a good performance.
    \item A neural network \textit{base encoder} $f(\cdot)$ that extracts representation vectors from augmented data examples.
    Our framework allows various choices of the network architecture without any constraints.
    We opt for simplicity and adopt the commonly used ResNet~\cite{he2016deep} to obtain $\bm h_i = f(\tilde{\bm x}_i) = \mathrm{ResNet}(\tilde{\bm x}_i)$ where $\bm h_i\in \mathbb{R}^d$ is the output after the average pooling layer.
    \item A small neural network \textit{projection head} $g(\cdot)$ that maps representations to the space where contrastive loss is applied.
    We use a MLP with one hidden layer to obtain $\bm z_i = g(\bm h_i)=W^{(2)}\sigma(W^{(1)}\bm h_i)$ where $\sigma$ is a ReLU non-linearity. As shown in section~\ref{sec:arch}, we find it beneficial
    to define the contrastive loss on $\bm z_i$'s rather than $\bm h_i$'s. 
    \item A \textit{contrastive loss function} defined for a contrastive prediction task. 
    Given a set $\{\tilde{\bm x}_k\}$ including a positive pair of examples $\tilde{\bm x}_i$ and $\tilde{\bm x}_j$, the \textit{contrastive prediction task} aims to identify $\tilde{\bm x}_j$ in $\{\tilde{\bm x}_k\}_{k\ne i}$ for a given $\tilde{\bm x}_i$.
\end{itemize}

We randomly sample a minibatch of $N$ examples and define the contrastive prediction task on pairs of augmented examples derived from the minibatch, resulting in $2N$ data points. We do not sample negative examples explicitly.
Instead, given a positive pair, similar to \cite{chen2017sampling}, we treat the other $2(N-1)$ augmented examples within a minibatch as negative examples.
Let $\mathrm{sim}(\bm u,\bm v) = \bm u^\top \bm v / \lVert\bm u\rVert \lVert\bm v\rVert$ denote the dot product between $\ell_2$ normalized $\bm u$ and $\bm v$ (i.e. cosine similarity). Then the loss function for a positive pair of examples $(i, j)$ is defined as
\begin{equation}
\label{eq:loss}
    \ell_{i,j} = -\log \frac{\exp(\mathrm{sim}(\bm z_i, \bm z_j)/\tau)}{\sum_{k=1}^{2N} \one{k \neq i}\exp(\mathrm{sim}(\bm z_i, \bm z_k)/\tau)}~,
\end{equation}
where $\one{k \neq i} \in \{ 0,  1\}$ is an indicator function evaluating to $1$ iff $k \neq i$ and $\tau$ denotes a temperature parameter. The final loss is computed across all positive pairs, both $(i, j)$ and $(j, i)$, in a mini-batch.
This loss has been used in previous work~\cite{sohn2016improved,wu2018unsupervised,oord2018representation}; for convenience, we term it \textit{NT-Xent} (the normalized temperature-scaled cross entropy loss).

Algorithm \ref{alg:main} summarizes the proposed method.

\begin{algorithm}[!t]
\caption{\label{alg:main} SimCLR's main learning algorithm.}
\begin{algorithmic}
    \STATE \textbf{input:} batch size $N$, constant $\tau$, structure of $f$, $g$, $\mathcal{T}$.
    \FOR{sampled minibatch $\{\bm x_k\}_{k=1}^N$}
    \STATE \textbf{for all} $k\in \{1, \ldots, N\}$ \textbf{do}
        \STATE $~~~~$draw two augmentation functions $t \!\sim\! \mathcal{T}$, $t' \!\sim\! \mathcal{T}$
        \STATE $~~~~$\textcolor{gray}{\# the first augmentation} \STATE $~~~~$$\tilde{\bm x}_{2k-1} = t(\bm x_k)$
        \STATE $~~~~$$\bm h_{2k-1} = f(\tilde{\bm x}_{2k-1})$  \textcolor{gray}{~~~~~~~~~~~~~~~~~~~~~~~~~~~~~~\# representation}
        \STATE $~~~~$$\bm z_{2k-1} = g({\bm h}_{2k-1})$  \textcolor{gray}{~~~~~~~~~~~~~~~~~~~~~~~~~~~~~~~~\# projection}
        \STATE $~~~~$\textcolor{gray}{\# the second augmentation} \STATE $~~~~$$\tilde{\bm x}_{2k} = t'(\bm x_k)$
        \STATE $~~~~$$\bm h_{2k} = f(\tilde{\bm x}_{2k})$      \textcolor{gray}{~~~~~~~~~~~~~~~~~~~~~~~~~~~~~~~~~~~~~~\# representation}
        \STATE $~~~~$$\bm z_{2k} = g({\bm h}_{2k})$      \textcolor{gray}{~~~~~~~~~~~~~~~~~~~~~~~~~~~~~~~~~~~~~~~~\# projection}
    \STATE \textbf{end for}
    \STATE \textbf{for all} $i\in\{1, \ldots, 2N\}$ and $j\in\{1, \dots, 2N\}$ \textbf{do}
    \STATE $~~~~$ $s_{i,j} = \bm z_i^\top \bm z_j / (\lVert\bm z_i\rVert \lVert\bm z_j\rVert)$ \textcolor{gray}{~~~~~~~~\# pairwise similarity}\\
    \STATE \textbf{end for}
    \STATE \textbf{define} $\ell(i, j)$ \textbf{as}~ $\ell(i, j) \!=\! -\log \frac{\exp(s_{i,j}/\tau)}{\sum_{k=1}^{2N} \one{k \neq i}\exp(s_{i, k}/\tau)}$ \\ \STATE $\mathcal{L} = \frac{1}{2N} \sum_{k=1}^N \left[ \ell(2k\!-\!1, 2k) + \ell(2k, 2k\!-\!1)\right]$
    \STATE update networks $f$ and $g$ to minimize $\mathcal{L}$
    \ENDFOR
    \STATE \textbf{return} encoder network $f(\cdot)$, and throw away $g(\cdot)$
\end{algorithmic}
\end{algorithm}

\subsection{Training with Large Batch Size}
To keep it simple, we do not train the model with a memory bank~\cite{wu2018unsupervised,he2019momentum}. Instead, we vary the training batch size $N$ from 256 to 8192. A batch size of 8192 gives us 16382 negative examples per positive pair from both augmentation views. Training with large batch size may be unstable when using standard SGD/Momentum with linear learning rate scaling~\cite{goyal2017accurate}. To stabilize the training, we use the LARS optimizer~\cite{you2017large} for all batch sizes.
We train our model with Cloud TPUs, using 32 to 128 cores depending on the batch size.\footnote{With 128 TPU v3 cores, it takes $\sim$1.5 hours to train our ResNet-50 with a batch size of 4096 for 100 epochs.}

\textbf{Global BN.} 
Standard ResNets use batch normalization~\cite{ioffe2015batch}. In distributed training with data parallelism, the BN mean and variance are typically aggregated locally per device. In our contrastive learning, as positive pairs are computed in the same device, the model can exploit the local information leakage to improve prediction accuracy without improving representations. We address this issue by aggregating BN mean and variance over all devices during the training. 
Other approaches include shuffling data examples across devices~\cite{he2019momentum}, or replacing BN with layer norm~\cite{henaff2019data}.

\subsection{Evaluation Protocol}
Here we lay out the protocol for our empirical studies, which aim to understand different design choices in our framework.

\textbf{Dataset and Metrics.} Most of our study for unsupervised pretraining (learning encoder network $f$ without labels) is done using the ImageNet ILSVRC-2012 dataset~\cite{russakovsky2015imagenet}. Some additional pretraining experiments on CIFAR-10~\cite{krizhevsky2009learning} can be found in Appendix~\ref{app:cifar}. We also test the pretrained results on a wide range of datasets for transfer learning.
To evaluate the learned representations, we follow the widely used linear evaluation protocol~\cite{zhang2016colorful,oord2018representation,bachman2019learning,kolesnikov2019revisiting}, where a linear classifier is trained on top of the frozen base network, and test accuracy is used as a proxy for representation quality. Beyond linear evaluation, we also compare against state-of-the-art on semi-supervised and transfer learning.

\textbf{Default setting.} Unless otherwise specified, for data augmentation we use random crop and resize (with random flip), color distortions, and Gaussian blur (for details, see Appendix~\ref{app:da}). We use ResNet-50 as the base encoder network, and a 2-layer MLP projection head to project the representation to a 128-dimensional latent space. As the loss, we use NT-Xent, optimized using LARS with learning rate of 4.8 ($=0.3\times\mathrm{BatchSize}/256$) and weight decay of $10^{-6}$. We train at batch size 4096 for 100 epochs.\footnote{Although max performance is not reached in 100 epochs, reasonable results are achieved, allowing fair and efficient ablations.} Furthermore, we use linear warmup for the first 10 epochs, and decay the learning rate with the cosine decay schedule without restarts~\cite{loshchilov2016sgdr}. \section{Data Augmentation for Contrastive Representation Learning}
\label{sec:da}

\begin{figure}[!t]
    \centering
    \begin{subfigure}[b]{0.22\textwidth}
    \centering
    \begin{tikzpicture}[scale=0.7]
        \draw[thick] (0,0) -- (4,0) -- (4,4) -- (0,4) -- (0,0);
        \draw[dashed] (1.1,1.1) -- (2.5,1.1) -- (2.5,2.5) -- (1.1,2.5) -- (1.1,1.1);
        \draw[dashed] (0.8,0.8) -- (3.5,0.8) -- (3.5,3.5) -- (0.8,3.5) -- (0.8,0.8);
        \node at (2.2, 1.4) {$A$};
        \node at (3.2, 1.4) {$B$};
    \end{tikzpicture}\caption{Global and local views.}
    \end{subfigure}
    ~
    \begin{subfigure}[b]{0.22\textwidth}
    \centering
    \begin{tikzpicture}[scale=0.7] 
        \draw[thick] (0,0) -- (4,0) -- (4,4) -- (0,4) -- (0,0);
        \draw[dashed] (0.5,0.5) -- (1.8,0.5) -- (1.8,1.8) -- (0.5,1.8) -- (0.5,0.5);
        \draw[dashed] (2,2) -- (3.8,2) -- (3.8,3.8) -- (2,3.8) -- (2,2);
        \node at (1.5, 0.9) {$C$};
        \node at (3.5, 2.3) {$D$};
    \end{tikzpicture}
    \caption{Adjacent views.}
    \end{subfigure}
    \vskip -0.4em
    \caption{\label{fig:trans_task}
    Solid rectangles are images, dashed rectangles are random crops. By randomly cropping images, we sample contrastive prediction tasks that include global to local view ($B\rightarrow A$) or adjacent view ($D\rightarrow C$) prediction.}
\end{figure}
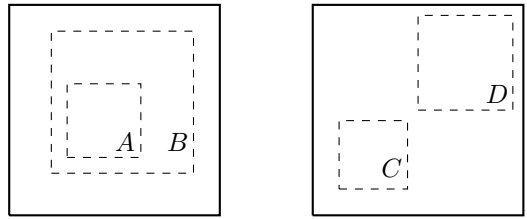

\begin{figure*}[!t]
\centering
\begin{subfigure}{.19\textwidth}
  \centering
  \includegraphics[width=0.9\linewidth]{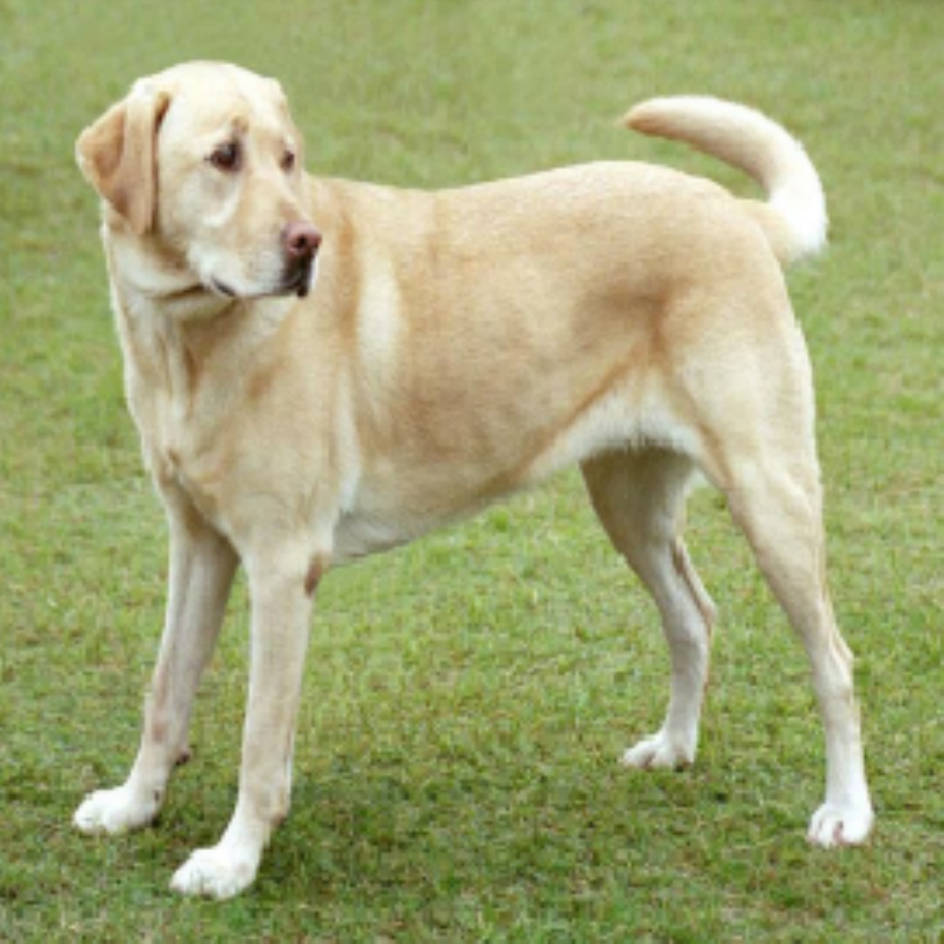}
  \caption{Original}
\end{subfigure}\begin{subfigure}{.19\textwidth}
  \centering
  \includegraphics[width=0.9\linewidth]{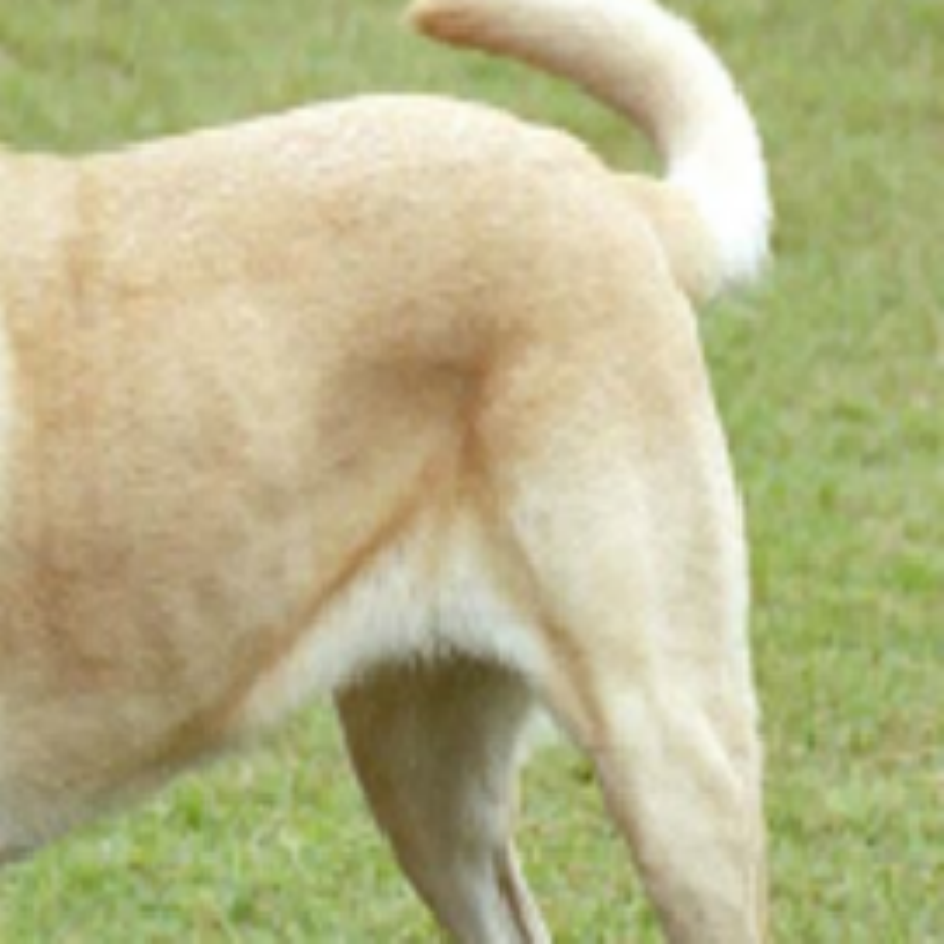}
  \caption{Crop and resize}
\end{subfigure}\begin{subfigure}{.19\textwidth}
  \centering
  \includegraphics[width=0.9\linewidth]{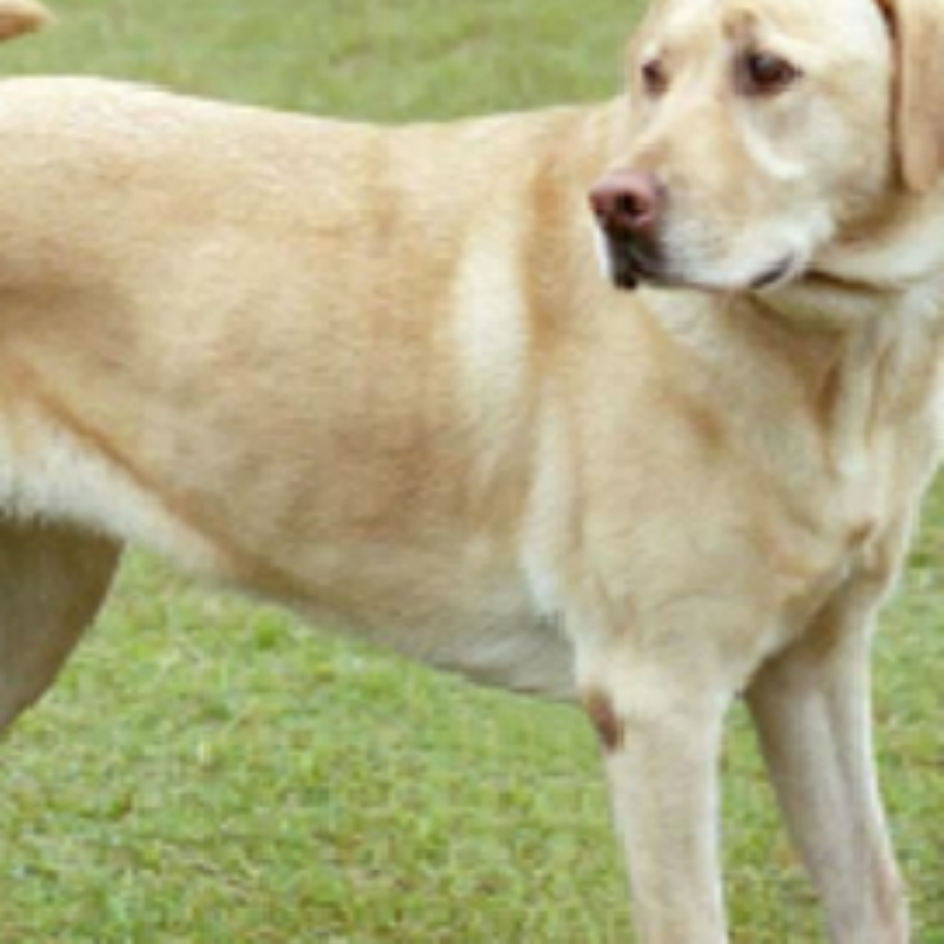}
  \caption{Crop, resize (and flip)}
\end{subfigure}\begin{subfigure}{.19\textwidth}
  \centering
  \includegraphics[width=0.9\linewidth]{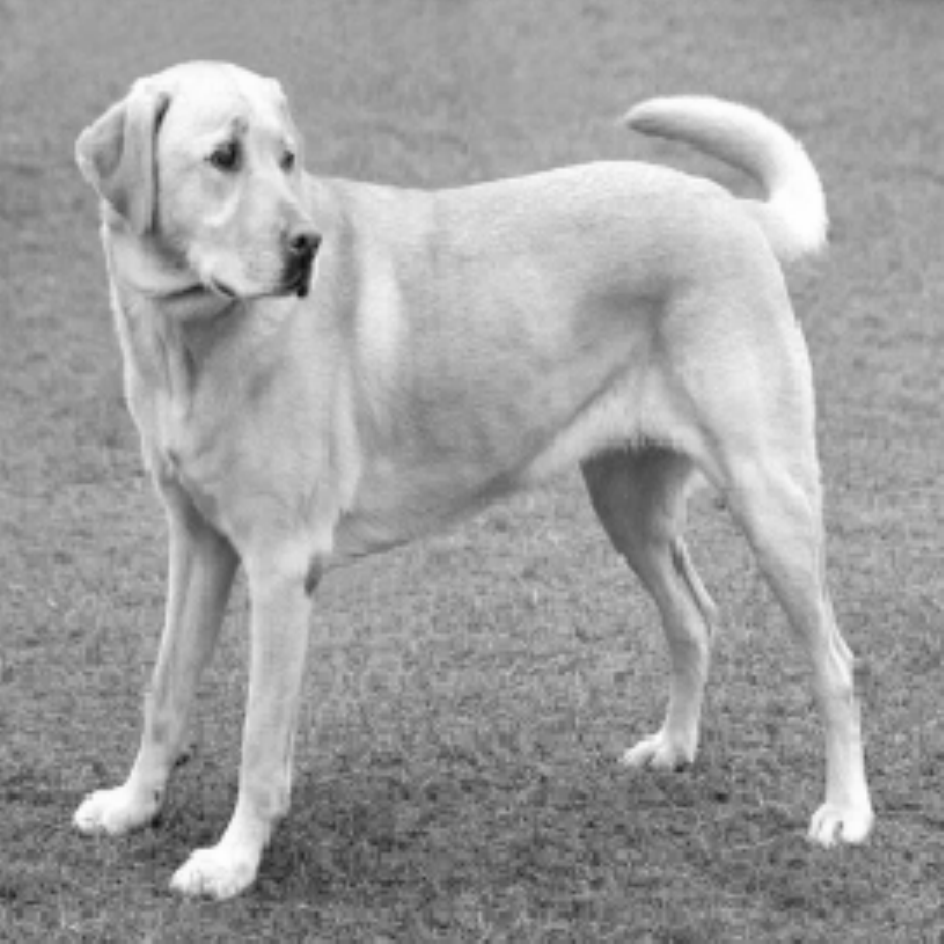}
  \caption{Color distort. (drop)}
\end{subfigure}\begin{subfigure}{.19\textwidth}
  \centering
  \includegraphics[width=0.9\linewidth]{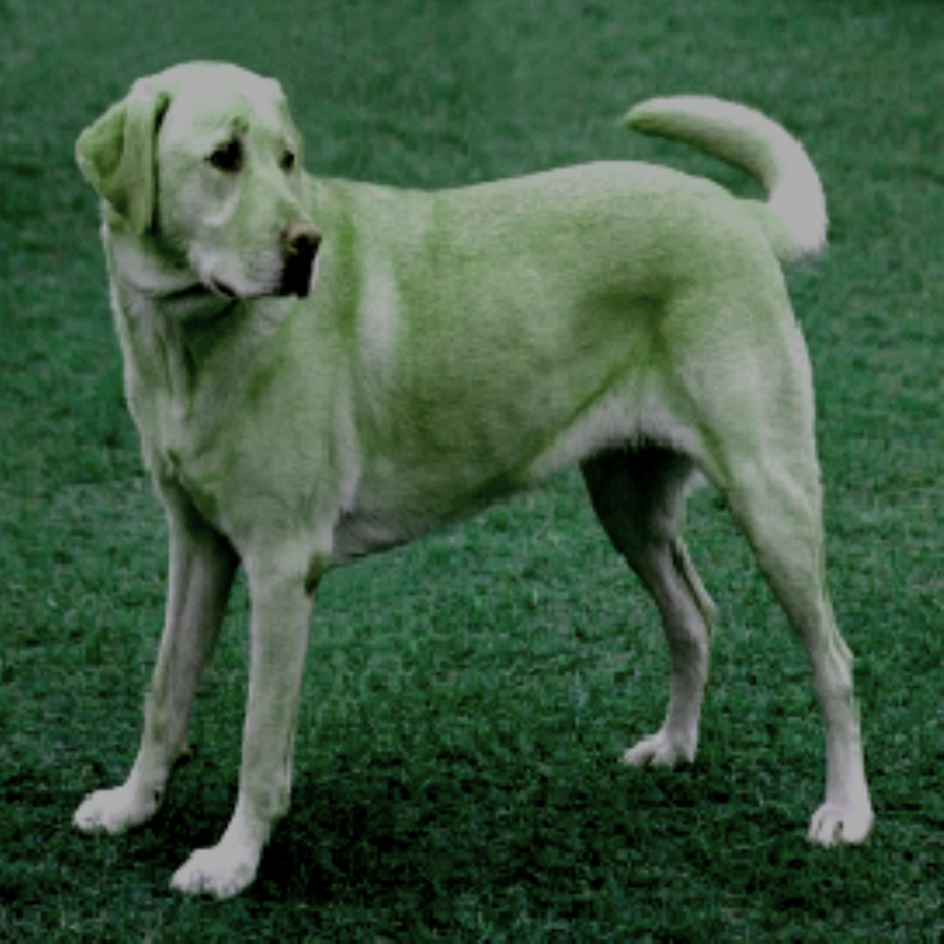}
  \caption{Color distort. (jitter)}
\end{subfigure}\\
\begin{subfigure}{.19\textwidth}
  \centering
  \includegraphics[width=0.9\linewidth]{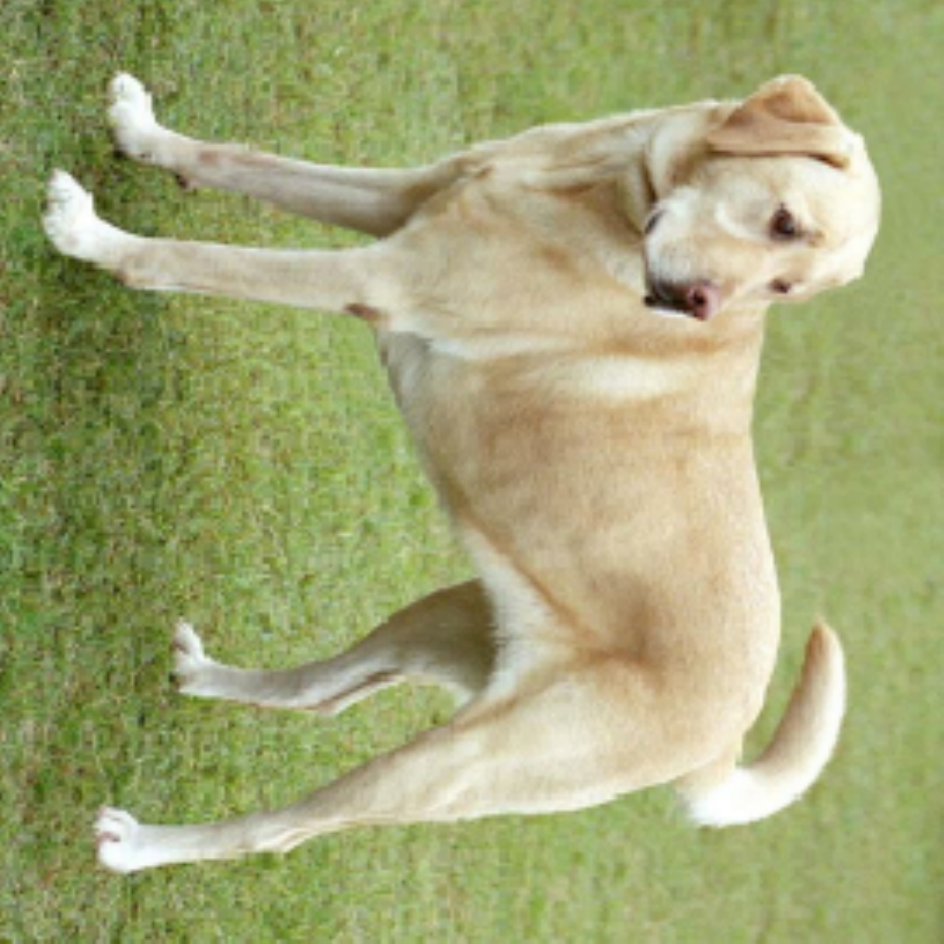}
  \caption{Rotate {\tiny$\{90\degree,180\degree,270\degree\}$}}
\end{subfigure}\begin{subfigure}{.19\textwidth}
  \centering
  \includegraphics[width=0.9\linewidth]{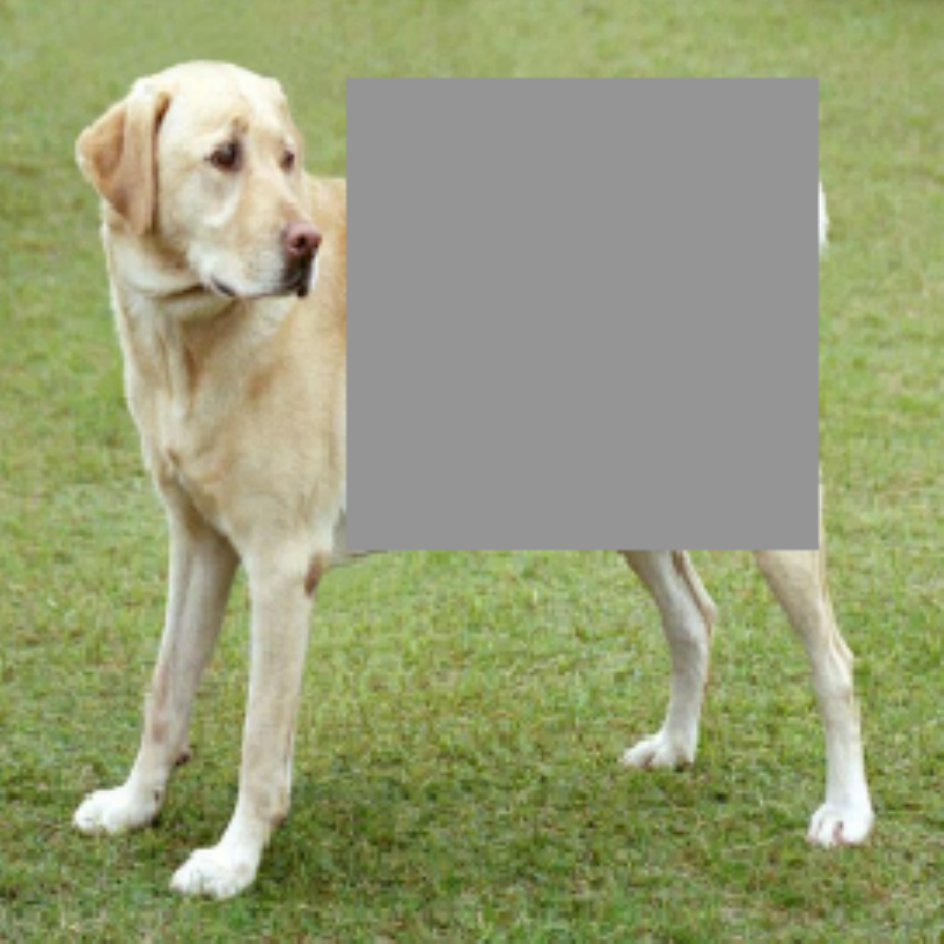}
  \caption{Cutout}
\end{subfigure}\begin{subfigure}{.19\textwidth}
  \centering
  \includegraphics[width=0.9\linewidth]{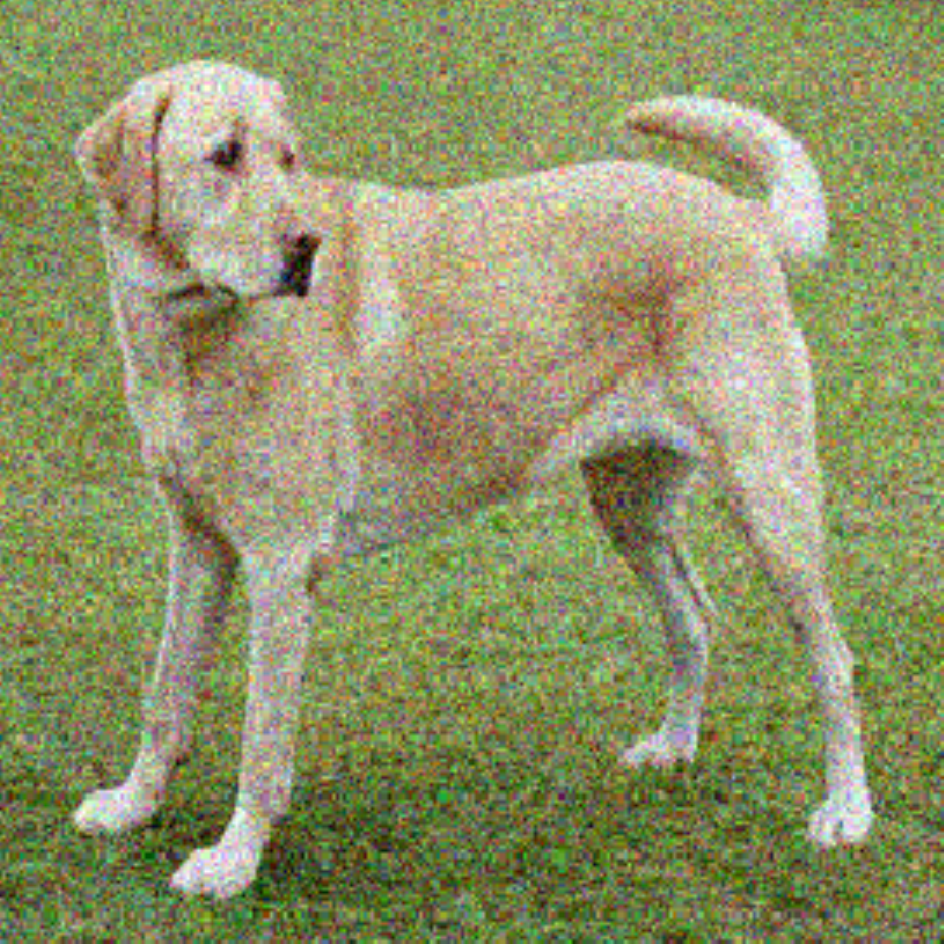}
  \caption{Gaussian noise}
\end{subfigure}\begin{subfigure}{.19\textwidth}
  \centering
  \includegraphics[width=0.9\linewidth]{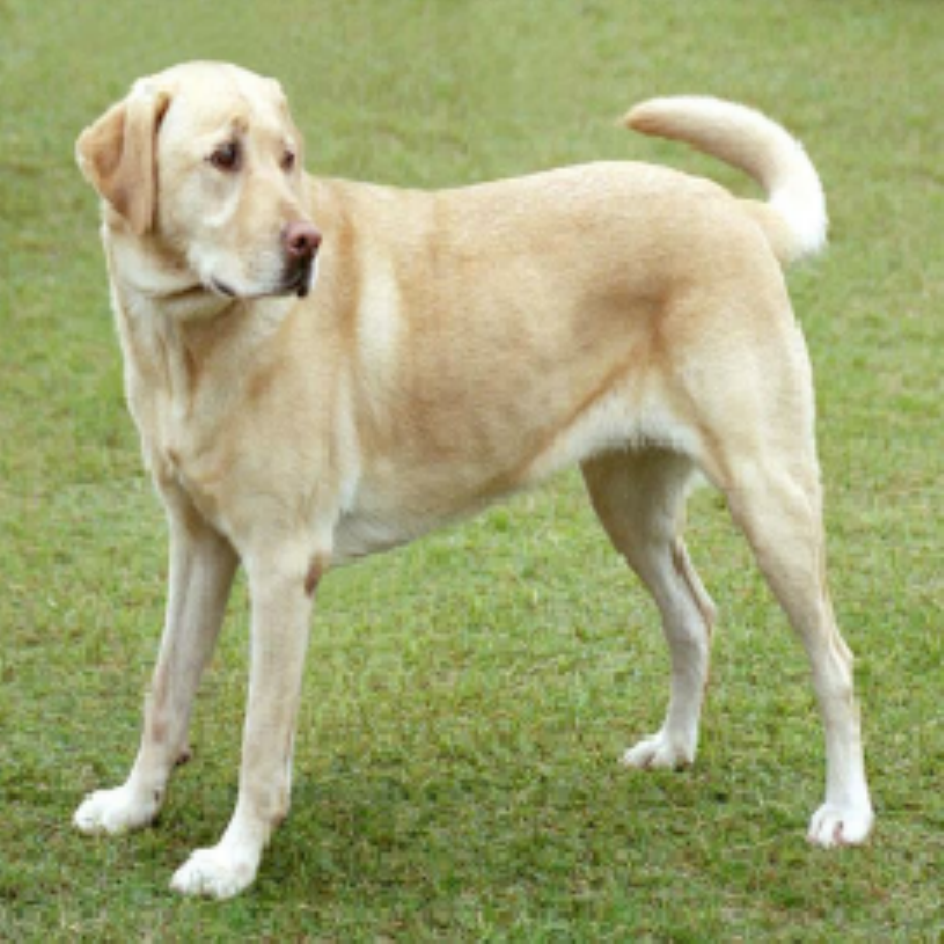}
  \caption{Gaussian blur}
\end{subfigure}\begin{subfigure}{.19\textwidth}
  \centering
  \includegraphics[width=0.9\linewidth]{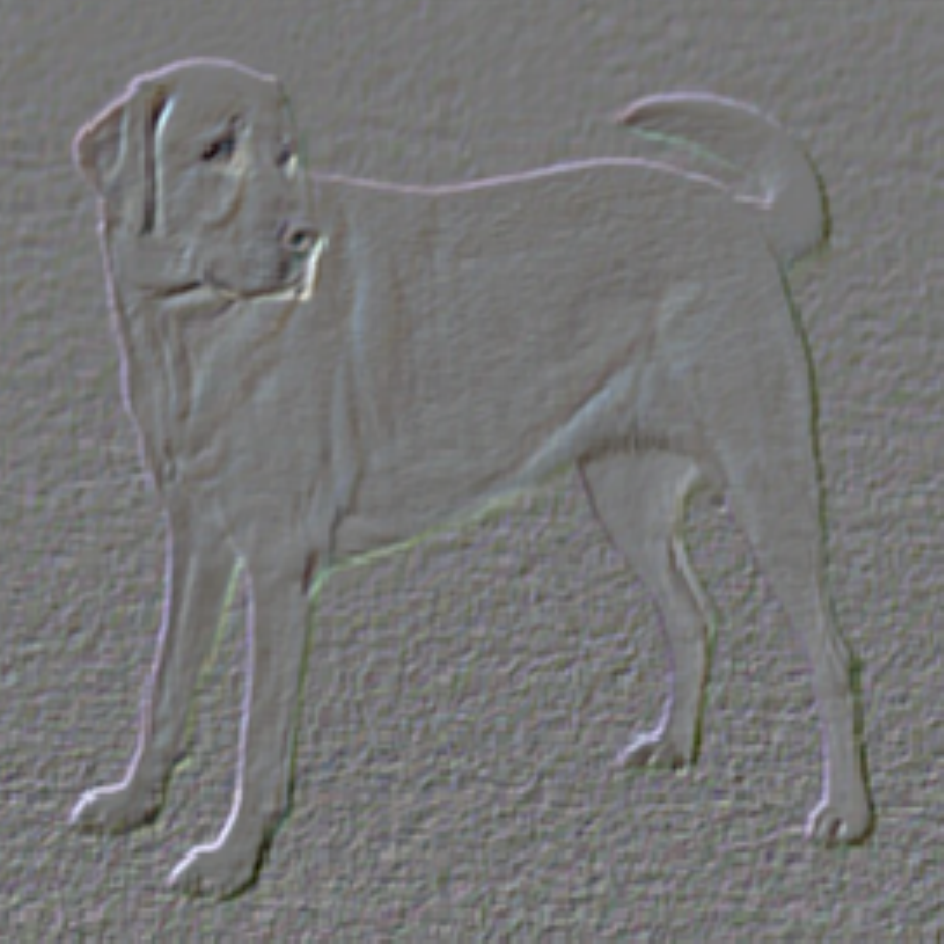}
  \caption{Sobel filtering}
\end{subfigure}\vskip -0.05in
\caption{Illustrations of the studied data augmentation operators. Each augmentation can transform data stochastically with some internal parameters (e.g. rotation degree, noise level). Note that we \textit{only} test these operators in ablation, the \textit{augmentation policy used to train our models} only includes \textit{random crop (with flip and resize)}, \textit{color distortion}, and \textit{Gaussian blur}. (Original image cc-by: Von.grzanka)}
\label{fig:data_aug}
\end{figure*}

\textbf{Data augmentation defines predictive tasks.}
While data augmentation has been widely used in both supervised and unsupervised representation learning~\cite{krizhevsky2012imagenet,henaff2019data,bachman2019learning}, it has not been considered as a systematic way to define the contrastive prediction task. Many existing approaches define contrastive prediction tasks by changing the architecture. For example, \citet{hjelm2018learning,bachman2019learning} achieve global-to-local view prediction via constraining the receptive field in the network architecture, whereas \citet{oord2018representation,henaff2019data} achieve neighboring view prediction via a fixed image splitting procedure and a context aggregation network. We show that this complexity can be avoided by performing simple \textit{random cropping} (with resizing) of target images, which creates a family of predictive tasks subsuming the above mentioned two, as shown in Figure~\ref{fig:trans_task}. This simple design choice conveniently decouples the predictive task from other components such as the neural network architecture. Broader contrastive prediction tasks can be defined by extending the family of augmentations and composing them stochastically.

\begin{figure}[!t]
    \centering
    \includegraphics[width=0.5\textwidth]{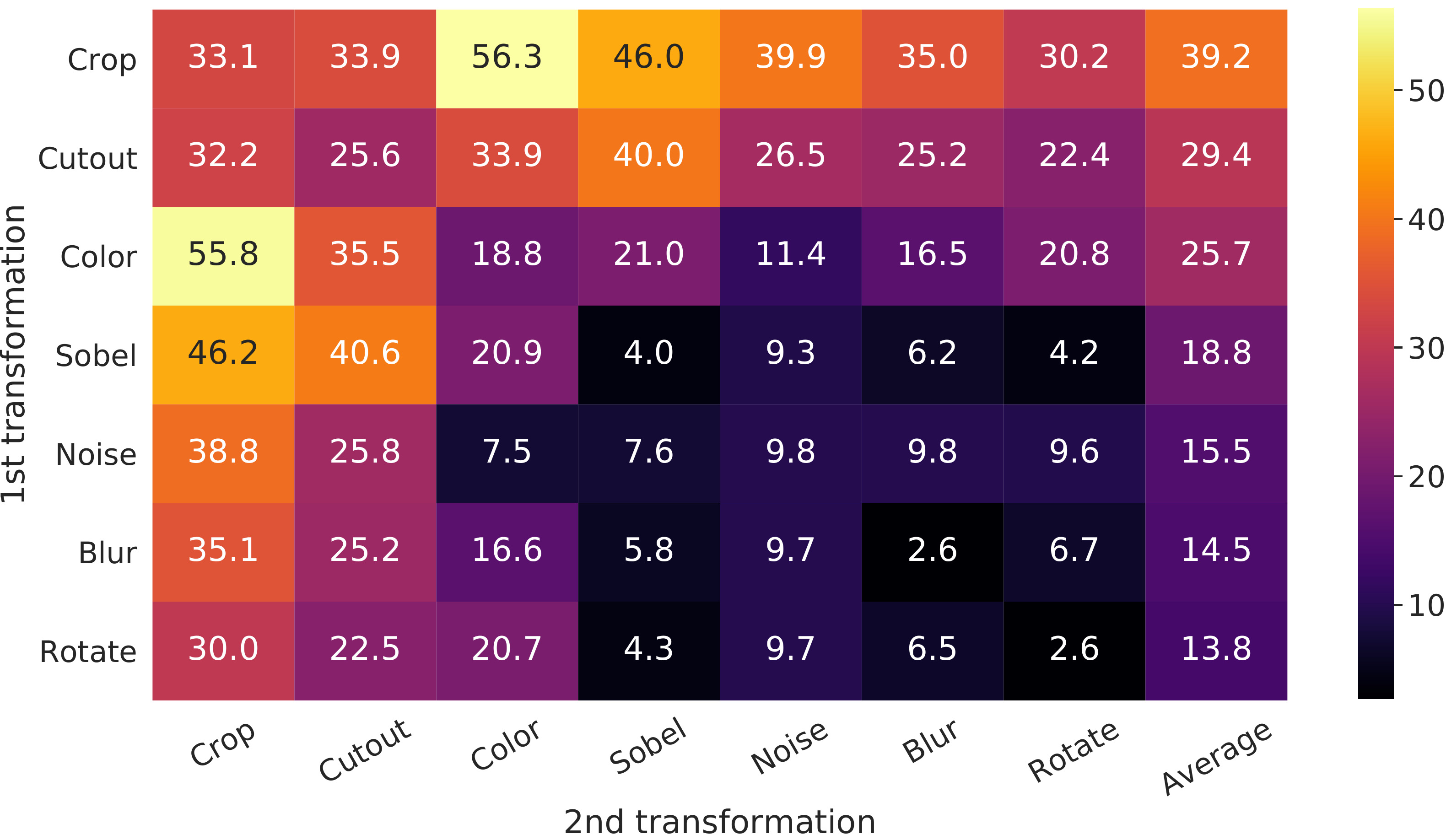}
    \vskip -0.4em
    \caption{\label{fig:da_pairwise}Linear evaluation (ImageNet top-1 accuracy) under individual or composition of data augmentations, applied only to one branch. For all columns but the last, diagonal entries correspond to single transformation, and off-diagonals correspond to composition of two transformations (applied sequentially). The last column reflects the average over the row.}
\end{figure}

\subsection{Composition of data augmentation operations is crucial for learning good representations}

To systematically study the impact of data augmentation, we consider several common augmentations here. One type of augmentation involves spatial/geometric transformation of data, such as cropping and resizing (with horizontal flipping), rotation~\cite{gidaris2018unsupervised} and cutout~\cite{devries2017improved}. The other type of augmentation involves appearance transformation, such as color distortion (including color dropping, brightness, contrast, saturation, hue)~\cite{howard2013some,szegedy2015going}, Gaussian blur, and Sobel filtering. Figure \ref{fig:data_aug} visualizes the augmentations that we study in this work.

To understand the effects of individual data augmentations and the importance of augmentation composition, we investigate the performance of our framework when applying augmentations individually or in pairs.
Since ImageNet images are of different sizes, we always apply crop and resize images~\cite{krizhevsky2012imagenet,szegedy2015going}, which makes it difficult to study other augmentations in the absence of cropping. To eliminate this confound, we consider an asymmetric data transformation setting for this ablation. Specifically, we always first randomly crop images and resize them to the same resolution, and we then apply the targeted transformation(s) \textit{only} to one branch of the framework in Figure~\ref{fig:framework}, while leaving the other branch as the identity (i.e. $t(\bm x_i) = \bm x_i$). Note that this asymmetric data augmentation hurts the performance. Nonetheless, this setup should not substantively change the impact of individual data augmentations or their compositions.

Figure \ref{fig:da_pairwise} shows linear evaluation results under individual and composition of transformations. We observe that \textit{no single transformation suffices to learn good representations}, even though the model can almost perfectly identify the positive pairs in the contrastive task. When composing augmentations, the contrastive prediction task becomes harder, but the quality of representation improves dramatically. Appendix~\ref{app:broader_augmentation} provides a further study on composing broader set of augmentations.

\begin{figure}[!t]
    \centering
    \begin{subfigure}{.24\textwidth}
      \centering
      \includegraphics[trim=0 160 0 0,clip,width=0.9\linewidth]{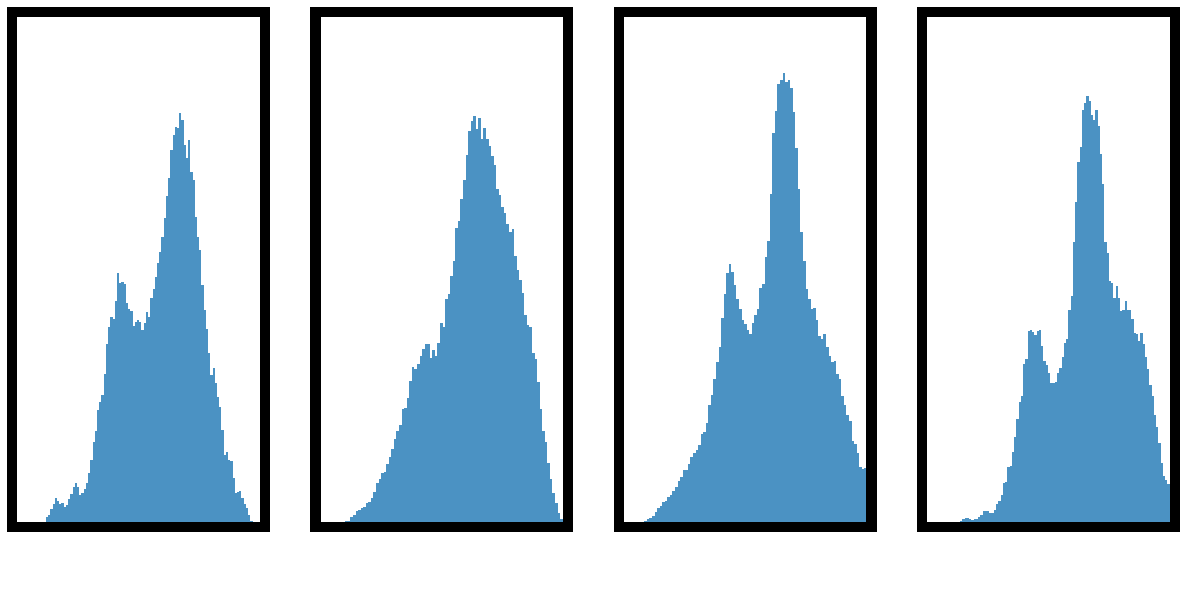}
    \end{subfigure}\begin{subfigure}{.24\textwidth}
      \centering
      \includegraphics[trim=0 160 0 0,clip,width=0.9\linewidth]{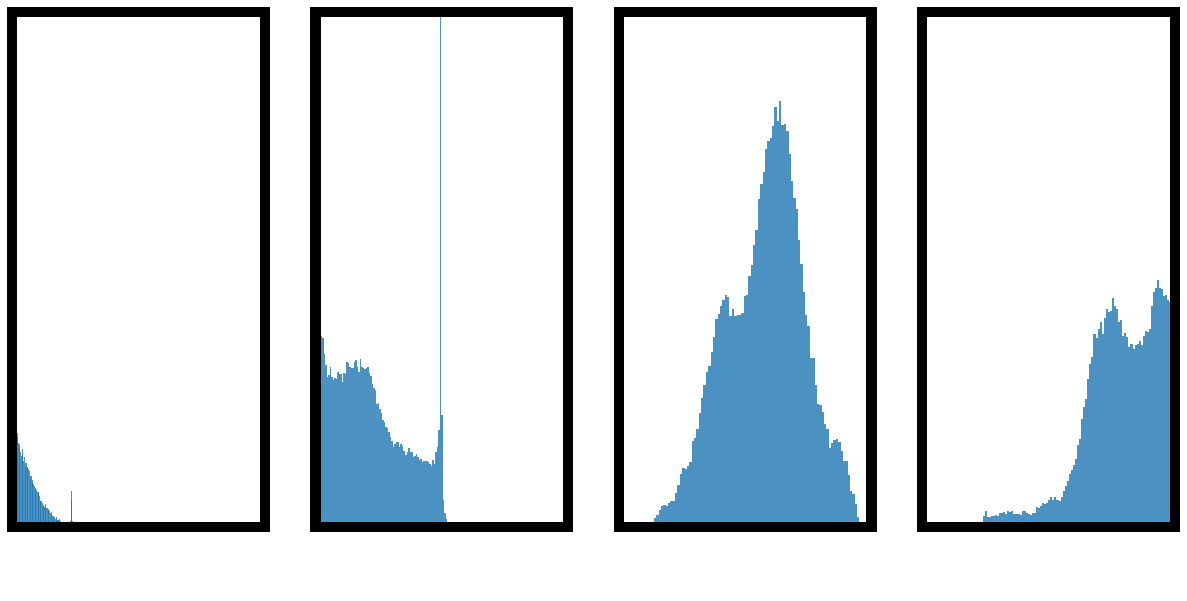}
    \end{subfigure}
    \begin{subfigure}{.24\textwidth}
      \centering
      \includegraphics[trim=0 170 0 0,clip,width=0.9\linewidth]{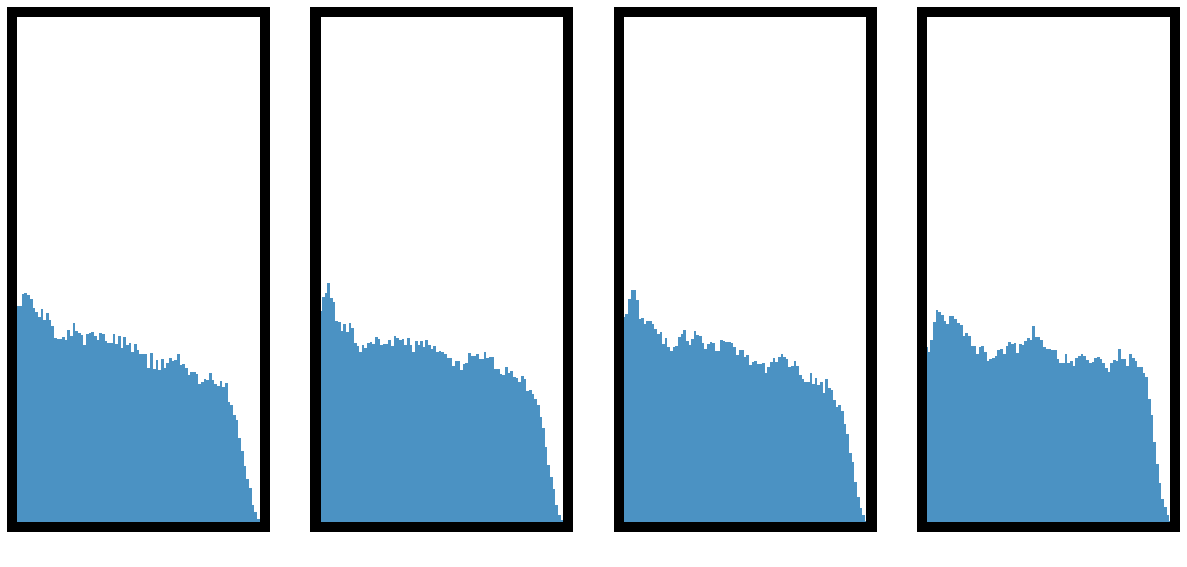}
      \caption{Without color distortion.}
    \end{subfigure}\begin{subfigure}{.24\textwidth}
      \centering
      \includegraphics[trim=0 170 0 0,clip,width=0.9\linewidth]{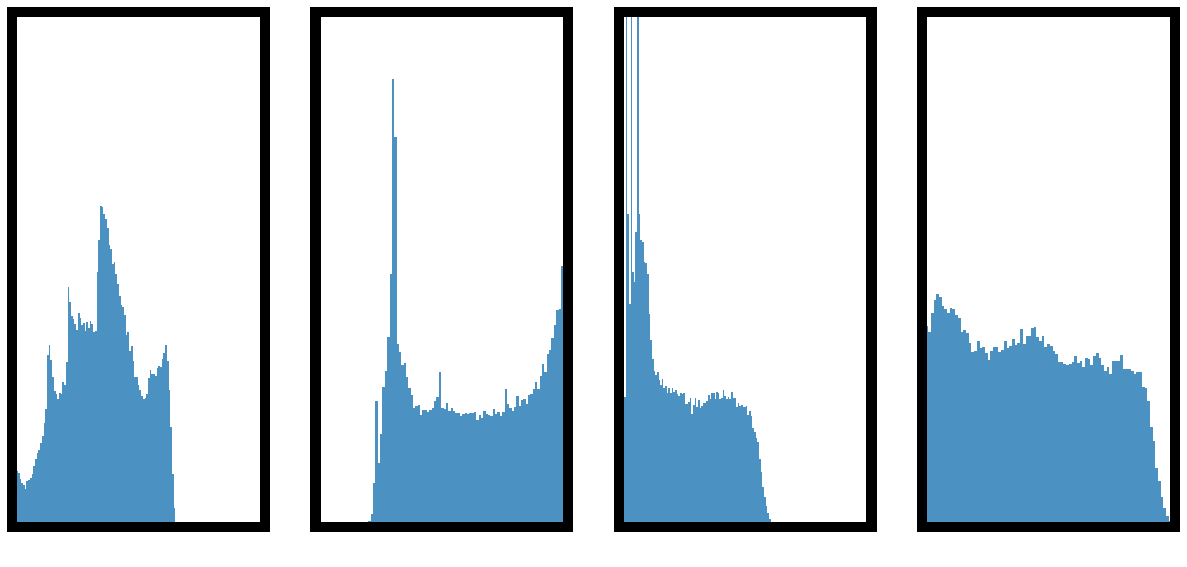}
      \caption{With color distortion.}
    \end{subfigure}
    \vskip -0.4em
    \caption{
Histograms of pixel intensities (over all channels) for
   different crops of two different images (i.e. two rows). The image for the first row is from Figure \ref{fig:data_aug}. All axes have the same range.}
    \label{fig:color_hist}
\end{figure}

One composition of augmentations stands out: random cropping and random color distortion. We conjecture that one serious issue when using only random cropping as data augmentation is that most patches from an image share a similar color distribution. Figure~\ref{fig:color_hist} shows that color histograms alone suffice to distinguish images. Neural nets may exploit this shortcut to solve the predictive task. Therefore, it is critical to compose cropping with color distortion in order to learn generalizable features. 

\subsection{Contrastive learning needs stronger data augmentation than supervised learning}

\begin{table}[!t]
    \centering
    \small
    \setlength{\tabcolsep}{5pt}
    \begin{tabular}{l|ccccc|c} \toprule
    &\multicolumn{5}{c|}{Color distortion strength} & \\
     Methods & 1/8 & 1/4 & 1/2 &1 & 1 (+Blur) & AutoAug\\ \midrule
     SimCLR &  59.6 & 61.0 & 62.6 & 63.2 & 64.5 & 61.1 \\
     Supervised &  77.0 & 76.7 & 76.5 & 75.7 & 75.4 & 77.1 \\ \bottomrule
    \end{tabular}
\caption{\label{fig:color_strength}Top-1 accuracy of unsupervised ResNet-50 using linear evaluation and supervised ResNet-50\footnotemark, under varied color distortion strength (see Appendix~\ref{app:da}) and other data transformations. Strength 1 (+Blur) is our default data augmentation policy.}
\end{table}
\footnotetext{Supervised models are trained for 90 epochs; longer training improves performance of stronger augmentation by $\sim0.5\%$.}

To further demonstrate the importance of the color augmentation, we adjust the strength of color augmentation as shown in Table \ref{fig:color_strength}. Stronger color augmentation substantially improves the linear evaluation of the learned unsupervised models. In this context, AutoAugment~\cite{cubuk2019autoaugment}, a sophisticated augmentation policy found using supervised learning, does not work better than simple cropping + (stronger) color distortion. When training supervised models with the same set of augmentations, we observe that stronger color augmentation does not improve or even hurts their performance. Thus, our experiments show that unsupervised contrastive learning benefits from stronger (color) data augmentation than supervised learning. Although previous work has reported that data augmentation is useful for self-supervised learning~\cite{doersch2015unsupervised,bachman2019learning,henaff2019data,asano2019critical}, we show that data augmentation that does not yield accuracy benefits for supervised learning can still help considerably with contrastive learning.

\section{Architectures for Encoder and Head}
\label{sec:arch}

\subsection{Unsupervised contrastive learning benefits (more) from bigger models}

\begin{figure}[!t]
\centering
\includegraphics[trim=0 0 0 0,clip,width=0.95\linewidth]{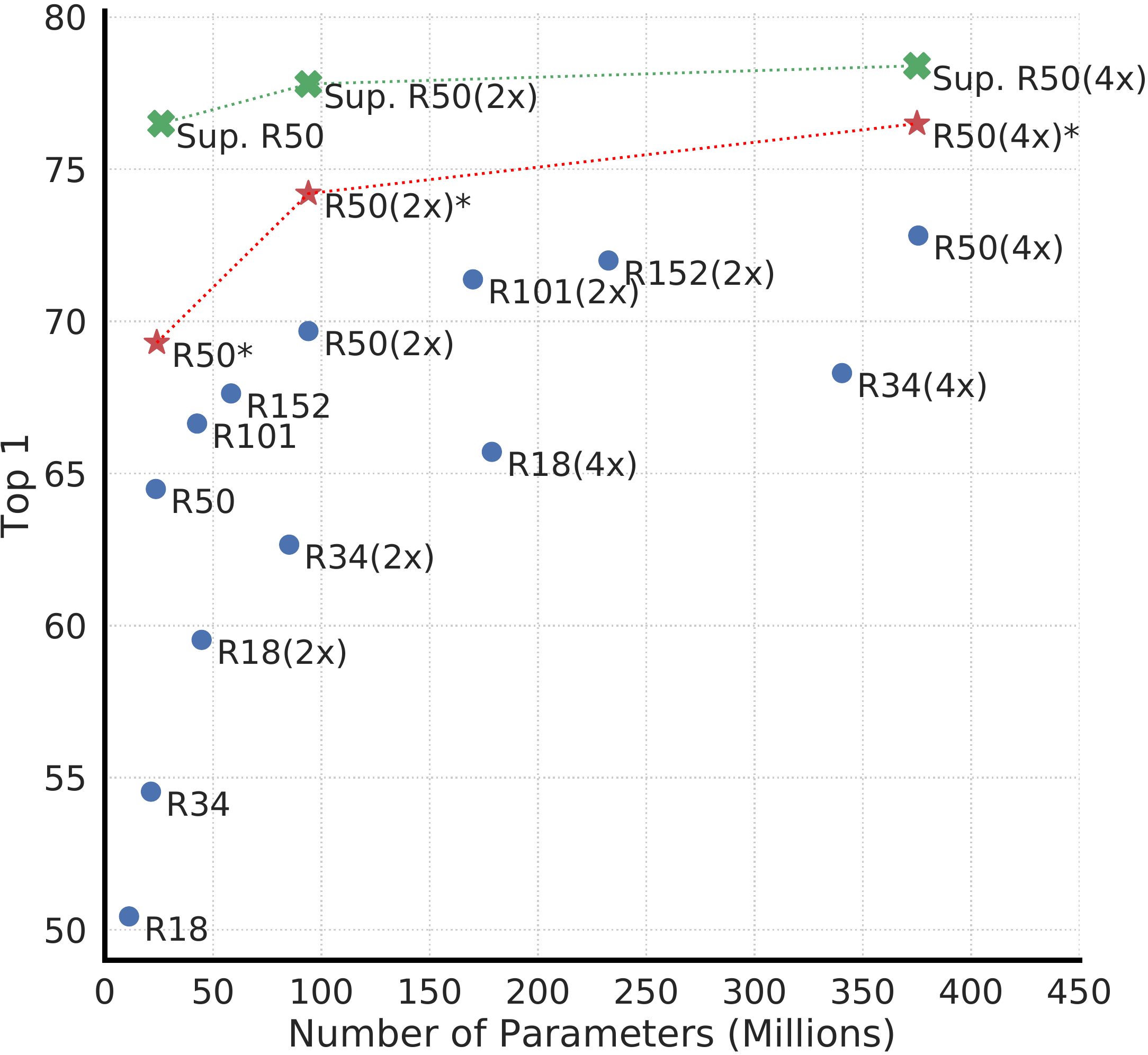}
\vskip -0.7em
\caption{\label{fig:arch}Linear evaluation of models with varied depth and width. Models in blue dots are ours trained for 100 epochs, models in red stars are ours trained for 1000 epochs, and models in green crosses are supervised ResNets trained for 90 epochs\footnotemark~\cite{he2016deep}.}
\end{figure}
\footnotetext{Training longer does not improve supervised ResNets (see Appendix~\ref{app:sup}).}

Figure \ref{fig:arch} shows, perhaps unsurprisingly, that increasing depth and width both improve performance. 
While similar findings hold for supervised learning~\cite{he2016deep}, we find the gap between supervised models and linear classifiers trained on unsupervised models shrinks as the model size increases, suggesting that \textit{unsupervised learning benefits more from bigger models than its supervised counterpart}.

\begin{table*}[!t]
    \small
    \centering
    \begin{tabular}{c|c|c} \toprule
         Name & Negative loss function&  Gradient w.r.t. $\bm u$\\ \midrule
         NT-Xent & $\bm u^T \bm v^+ /\tau - \log\sum_{\bm v\in\{\bm v^+, \bm v^-\}}\exp(\bm u^T \bm v/\tau)$ &  $(1-\frac{\exp(\bm u^T \bm v^+ /\tau)}{Z(\bm u)})/\tau \bm v^+ - \sum_{\bm v^-}\frac{\exp(\bm u^T \bm v^-/\tau)}{Z(\bm u)}/\tau \bm v^-$ \\ [6pt]
         NT-Logistic &$\log\sigma(\bm u^T \bm v^+ /\tau) + \log\sigma(-\bm u^T \bm v^-/\tau)$ & $(\sigma(-\bm u^T \bm v^+ /\tau))/\tau \bm v^+ - \sigma(\bm u^T \bm v^-/\tau)/\tau \bm v^-$ \\  [6pt]
         Margin Triplet &$-\max(\bm u^T \bm v^- - \bm u^T \bm v^+ + m, 0)$ &  $\bm v^+ - \bm v^- \text{ if } \bm u^T \bm v^+ - \bm u^T \bm v^- < m \text{ else } \bm 0$  \\  [6pt]
\bottomrule
    \end{tabular}
\caption{Negative loss functions and their gradients. All input vectors, i.e. $\bm u, \bm v^+, \bm v^-$, are $\ell_2$ normalized.
    NT-Xent is an abbreviation for ``Normalized Temperature-scaled Cross Entropy''. Different loss functions impose different weightings of positive and negative examples.}
    \label{tab:loss}
\end{table*}

\subsection{A nonlinear projection head improves the representation quality of the layer before it}

\begin{figure}[!t]
    \centering
    \vskip -0.5em
    \includegraphics[width=.35\textwidth]{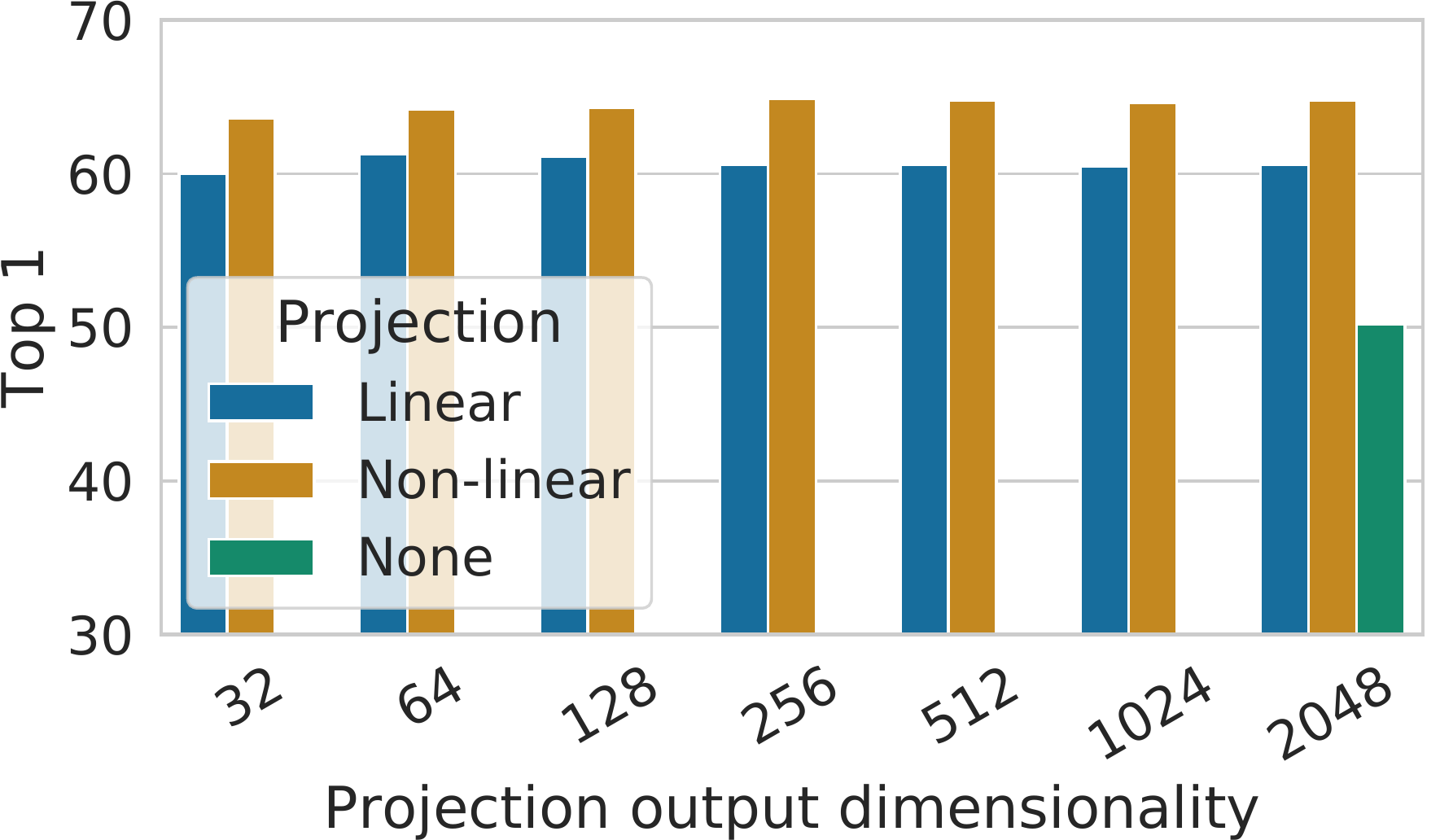}
    \vskip -0.5em
    \caption{\label{fig:critic_proj}Linear evaluation of representations with different projection heads $g(\cdot)$
    and various dimensions of $\bm z = g(\bm h)$. The representation $\bm h$ (before projection) is 2048-dimensional here.}\vskip -0.5em
\end{figure}

We then study the importance of including a projection head, i.e. $g(\bm h)$. Figure \ref{fig:critic_proj} shows linear evaluation results using three different architecture for the head: (1) identity mapping; (2) linear projection, as used by several previous approaches~\cite{wu2018unsupervised}; and (3) the default nonlinear projection with one additional hidden layer (and ReLU activation), similar to \citet{bachman2019learning}. We observe that a nonlinear projection is better than a linear projection (+3\%), and much better than no projection (>10\%). When a projection head is used, similar results are observed regardless of output dimension. Furthermore, even when nonlinear projection is used, the layer before the projection head, $\bm h$, is still much better (>10\%) than the layer after, $\bm z=g(\bm h)$, which shows that \textit{the hidden layer before the projection head is a better representation than the layer after}.

We conjecture that the importance of using the representation before the nonlinear projection is due to loss of information induced by the contrastive loss. In particular, $\bm z=g(\bm h)$ is trained to be invariant to data transformation. Thus, $g$ can remove information that may be useful for the downstream task, such as the color or orientation of objects. By leveraging the nonlinear transformation $g(\cdot)$, more information can be formed and maintained in $\bm h$. To verify this hypothesis, we conduct experiments that use either $\bm h$ or $g(\bm h)$ to learn to predict the transformation applied during the pretraining. Here we set $g(h)=W^{(2)}\sigma(W^{(1)}h)$, with the same input and output dimensionality (i.e. 2048). Table \ref{tab:critic_invariance} shows $\bm h$ contains much more information about the transformation applied, while $g(\bm h)$ loses information. Further analysis can be found in Appendix~\ref{app:understand_head}.

\begin{table}[!t]
\centering
\small
\begin{tabular}{lccc}
\toprule
\multirow{2}{*}{What to predict?} &\multirow{2}{*}{Random guess} & \multicolumn{2}{c}{Representation} \\
    &     & $\bm h$      & $g(\bm h)$ \\ \midrule
Color vs grayscale     & 80    & 99.3 & 97.4\\
Rotation    & 25   & 67.6    & 25.6   \\ 
Orig. vs corrupted  & 50  & 99.5   & 59.6   \\
Orig. vs Sobel filtered & 50 & 96.6   & 56.3   \\
\bottomrule
\end{tabular}
\vskip -0.4em
\caption{\label{tab:critic_invariance}Accuracy of training additional MLPs on different representations to predict the transformation applied. Other than crop and color augmentation, we additionally and independently add rotation (one of $\{0\degree, 90\degree, 180\degree, 270\degree\}$), Gaussian noise, and Sobel filtering transformation during the pretraining for the last three rows. Both $\bm h$ and $g(\bm h)$ are of the same dimensionality, i.e. 2048.}
\end{table}

\section{Loss Functions and Batch Size}

\subsection{Normalized cross entropy loss with adjustable temperature works better than alternatives}

We compare the NT-Xent loss against other commonly used contrastive loss functions, such as logistic loss~\cite{mikolov2013efficient}, and margin loss~\cite{schroff2015facenet}. Table \ref{tab:loss} shows the objective function as well as the gradient to the input of the loss function. Looking at the gradient, we observe 1) $\ell_2$ normalization (i.e. cosine similarity) along with temperature effectively weights different examples, and an appropriate temperature can help the model learn from hard negatives; and 2) unlike cross-entropy, other objective functions do not weigh the negatives by their relative hardness. As a result, one must apply semi-hard negative mining~\cite{schroff2015facenet} for these loss functions: instead of computing the gradient over all loss terms, one can compute the gradient using semi-hard negative terms (\textit{i.e.}, those that are within the loss margin and closest in distance, but farther than positive examples).

To make the comparisons fair, we use the same $\ell_2$ normalization for all loss functions, and we tune the hyperparameters, and report their best results.\footnote{Details can be found in Appendix~\ref{app:loss_tuning}. For simplicity, we only consider the negatives from one augmentation view.
} Table \ref{tab:loss_comp} shows that, while (semi-hard) negative mining helps, the best result is still much worse than our default NT-Xent loss.

We next test the importance of the $\ell_2$ normalization (i.e. cosine similarity vs dot product) and temperature $\tau$ in our default NT-Xent loss. Table \ref{tab:loss_norm} shows that without normalization and proper temperature scaling, performance is significantly worse. Without $\ell_2$ normalization, the contrastive task accuracy is higher, but the resulting representation is worse under linear evaluation.

\begin{table}[!t]
    \small
    \centering
    \setlength{\tabcolsep}{5pt}
    \begin{tabular}{ccccc} \toprule
    Margin	&NT-Logi.	&Margin (sh)	&NT-Logi.(sh)	&NT-Xent \\ \midrule 
    50.9	&51.6	&57.5	&57.9 & 63.9\\\bottomrule 
    \end{tabular}
    \vskip -0.4em
    \caption{Linear evaluation (top-1) for models trained with different loss functions. ``sh'' means using semi-hard negative mining.}
    \label{tab:loss_comp}
\end{table}

\begin{table}[!t]
    \small
    \centering
    \begin{tabular}{cc|cc|c} \toprule
    $\ell_2$ norm? &$\tau$ & Entropy  & Contrastive acc. & Top 1\\ \midrule
    \multirow{4}{*}{Yes}	& 0.05& 1.0 & 90.5	&{59.7}	\\
    & 0.1	& 4.5 & 87.8	&{64.4}\\
    & 0.5	& 8.2 & 68.2	&{60.7}\\
    & 1	&8.3 & 59.1	&58.0\\ \midrule
    \multirow{2}{*}{No} & 10	& 0.5 & 91.7	&{57.2} \\
    & 100	& 0.5 &92.1	&57.0 \\ \bottomrule
    \end{tabular}
    \vskip -0.4em
    \caption{Linear evaluation for models trained with different choices of $\ell_2$ norm and temperature $\tau$ for NT-Xent loss. The contrastive distribution is over 4096 examples.}
\label{tab:loss_norm}
\end{table}

\subsection{Contrastive learning benefits (more) from larger batch sizes and longer training}

\begin{figure}[!t]
\centering
\vskip -0.3em
\includegraphics[width=\linewidth]{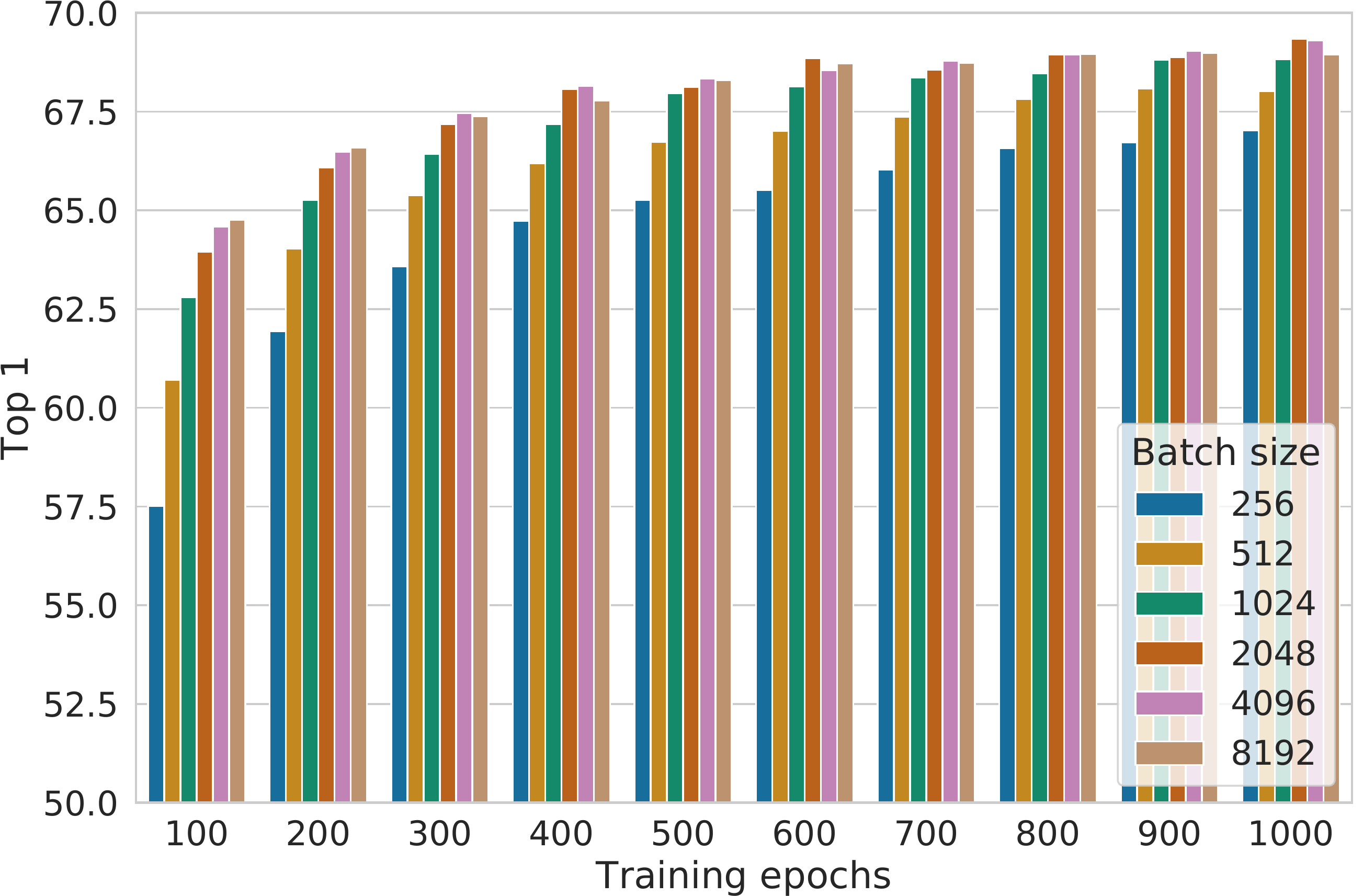}
\vskip -0.6em
\caption{\label{fig:bsstep_top1}Linear evaluation models (ResNet-50) trained with different batch size and epochs. Each bar is a single run from scratch.\footnotemark}
\end{figure}
\footnotetext{A linear learning rate scaling is used here. Figure~\ref{tab:sqrt_vs_linear_scaling} shows using a square root learning rate scaling can improve performance of ones with small batch sizes.}

Figure \ref{fig:bsstep_top1} shows the impact of batch size when models are trained for different numbers of epochs. We find that, when the number of training epochs is small (e.g. 100 epochs), larger batch sizes have a significant advantage over the smaller ones. With more training steps/epochs, the gaps between different batch sizes decrease or disappear, provided the batches are randomly resampled. 
In contrast to supervised learning~\cite{goyal2017accurate}, in contrastive learning, larger batch sizes provide more negative examples, facilitating convergence (i.e. taking fewer epochs and steps for a given accuracy). Training longer also provides more negative examples, improving the results. In Appendix~\ref{app:bsz_step}, results with even longer training steps are provided.

\begin{table}[!t]
\small
\centering
\setlength{\tabcolsep}{5pt}
\begin{tabular}{ll@{\hspace{0cm}}c@{\hspace{0.3cm}}cc@{\hspace{0.2cm}}}
\toprule
Method & Architecture         & Param (M) & Top 1 & Top 5 \\ \midrule
\multicolumn{5}{l}{\textit{Methods using ResNet-50:}}                   \\
Local Agg.   & ResNet-50            & 24             & 60.2  & -     \\
MoCo   & ResNet-50            & 24             & 60.6  & -     \\
PIRL   & ResNet-50            & 24             & 63.6  & -  \\
CPC v2 & ResNet-50            & 24             & 63.8  & 85.3  \\
SimCLR (ours)  & ResNet-50            & 24             & \textbf{69.3}    & \textbf{89.0}    \\ \midrule
\multicolumn{5}{l}{\textit{Methods using other architectures:}}         \\ 
Rotation  & RevNet-50 ($4\times$)  & 86            & 55.4  &    -   \\ 
BigBiGAN  & RevNet-50 ($4\times$)  & 86            & 61.3  &    81.9   \\ 
AMDIM  & Custom-ResNet           & 626            & 68.1  &    -   \\
CMC   & ResNet-50 ($2\times$) & 188            & 68.4  &    88.2   \\
MoCo   & ResNet-50 ($4\times$) & 375            & 68.6  &    -   \\
CPC v2 & ResNet-161 ($*$)          & 305            & 71.5  & 90.1  \\
SimCLR (ours)   & ResNet-50 ($2\times$) & 94             & 74.2    & 92.0    \\
SimCLR (ours)   & ResNet-50 ($4\times$) & 375            & \textbf{76.5}    & \textbf{93.2}   \\ 
\bottomrule
\end{tabular}
\vskip -0.4em
\caption{\label{tab:linear_sota}ImageNet accuracies of linear classifiers trained on representations learned with different self-supervised methods.}
\end{table}

\begin{table}[!t]
\small
\centering
\begin{tabular}{@{\hspace{.2cm}}l@{\hspace{.3cm}}lcc@{\hspace{.2cm}}}
\toprule
\multirow{3}{*}{Method} & \multirow{3}{*}{Architecture} & \multicolumn{2}{c}{Label fraction} \\
& & 1\% & 10\% \\
& & \multicolumn{2}{c}{Top 5} \\
\midrule
Supervised baseline      & ResNet-50  &  48.4 & 80.4 \\ \midrule
\multicolumn{4}{l}{\textit{Methods using other label-propagation:}}         \\ 
Pseudo-label          & ResNet-50 &  51.6 & 82.4\\
VAT+Entropy Min.          & ResNet-50 &  47.0 & 83.4 \\
UDA (w. RandAug)                        & ResNet-50 &  - & 88.5 \\
FixMatch (w. RandAug)         & ResNet-50 & - & 89.1 \\
S4L (Rot+VAT+En. M.)          & ResNet-50 (4$\times$) &  - & 91.2 \\ \midrule
\multicolumn{4}{l}{\textit{Methods using representation learning only:}}         \\ 
InstDisc  & ResNet-50 & 39.2 & 77.4\\
BigBiGAN & RevNet-50 ($4\times$)             & 55.2  & 78.8  \\
PIRL & ResNet-50             & 57.2  & 83.8  \\
CPC v2 & ResNet-161($*$)            & 77.9  & 91.2  \\
SimCLR (ours)  & ResNet-50        & 75.5
   & 	87.8   \\
SimCLR (ours)  & ResNet-50 ($2\times$)           & 83.0  &  91.2  \\
SimCLR (ours)  & ResNet-50 ($4\times$)           & \textbf{85.8}
   &  \textbf{92.6} \\
\bottomrule
\end{tabular}
\vskip -0.4em
\caption{\label{tab:data_efficient_sota}ImageNet accuracy of models trained with few labels.}
\end{table}

\begin{table*}[tb]
\footnotesize
\centering
\setlength{\tabcolsep}{3pt}
\begin{tabular}{lcccccccccccc}
\toprule
{} &  Food &  CIFAR10 &  CIFAR100 &  Birdsnap &  SUN397 &  Cars &  Aircraft &  VOC2007 &  DTD &  Pets &  Caltech-101 &  Flowers \\
\midrule
\multicolumn{5}{l}{\textit{Linear evaluation:}}\\
SimCLR (ours) &     \textbf{76.9} &     \textbf{95.3} &      80.2 &      48.4 &    \textbf{65.9} &           60.0 &           61.2 &     \textbf{84.2} &        \textbf{78.9} &         89.2 &               \textbf{93.9} &               \textbf{95.0} \\
Supervised &     75.2 &     \textbf{95.7} &      \textbf{81.2} &      \textbf{56.4} &    64.9 &           \textbf{68.8} &           \textbf{63.8} &     83.8 &        \textbf{78.7} &         \textbf{92.3} &               \textbf{94.1} &               94.2 \\
\midrule
\multicolumn{5}{l}{\textit{Fine-tuned:}}\\
SimCLR (ours) &     \textbf{89.4} &     \textbf{98.6} &      \textbf{89.0} &      \textbf{78.2} &    \textbf{68.1} &           \textbf{92.1} &           \textbf{87.0} &     \textbf{86.6} &        \textbf{77.8} &         92.1 &               \textbf{94.1} &               97.6 \\
Supervised &     88.7 &     98.3 &      \textbf{88.7} &      \textbf{77.8} &    67.0 &           91.4 &           \textbf{88.0} &     86.5 &        \textbf{78.8} &         \textbf{93.2} &               \textbf{94.2} &               \textbf{98.0} \\
Random init &     88.3 &     96.0 &      81.9 &      \textbf{77.0} &    53.7 &           91.3 &           84.8 &     69.4 &        64.1 &         82.7 &               72.5 &               92.5 \\

\bottomrule
\end{tabular}
\caption{Comparison of transfer learning performance of our self-supervised approach with supervised baselines across 12 natural image classification datasets, for ResNet-50 $(4\times)$ models pretrained on ImageNet. Results not significantly worse than the best ($p > 0.05$, permutation test) are shown in bold. See Appendix~\ref{app:transfer_learning} for experimental details and results with standard ResNet-50.}
\label{tab:transfer_learning_resnet_4x}
\end{table*}

\section{Comparison with State-of-the-art}
In this subsection, similar to~\citet{kolesnikov2019revisiting,he2019momentum}, we use ResNet-50 in 3 different hidden layer widths (width multipliers of $1\times$, $2\times$, and $4\times$). For better convergence, our models here are trained for 1000 epochs.

\textbf{Linear evaluation.} Table \ref{tab:linear_sota} compares our results with previous approaches~\cite{zhuang2019local,he2019momentum,misra2019self,henaff2019data,kolesnikov2019revisiting,donahue2019large,bachman2019learning,tian2019contrastive} in the linear evaluation setting (see Appendix~\ref{app:result_linear}). Table \ref{fig:linear_sota} shows more numerical comparisons among different methods. We are able to use standard networks to obtain substantially better results compared to previous methods that require specifically designed architectures. The best result obtained with our ResNet-50 ($4\times$) can match the supervised pretrained ResNet-50.

\textbf{Semi-supervised learning.} We follow~\citet{zhai2019s4l} and sample 1\% or 10\% of the labeled ILSVRC-12 training datasets in a class-balanced way ($\sim$12.8 and $\sim$128 images per class respectively).~\footnote{The details of sampling and exact subsets can be found in \href{https://www.tensorflow.org/datasets/catalog/imagenet2012\_subset}{https://www.tensorflow.org/datasets/catalog/imagenet2012\_subset}.} We simply fine-tune the whole base network on the labeled data without regularization (see Appendix~\ref{app:result_semi}). Table \ref{tab:data_efficient_sota} shows the comparisons of our results against recent methods~\cite{zhai2019s4l,xie2019unsupervised,sohn2020fixmatch,wu2018unsupervised,donahue2019large,misra2019self,henaff2019data}. The supervised baseline from~\cite{zhai2019s4l} is strong due to intensive search of hyper-parameters (including augmentation). Again, our approach significantly improves over state-of-the-art with both 1\% and 10\% of the labels. Interestingly, fine-tuning our pretrained ResNet-50 (2$\times, 4\times$) on \textit{full} ImageNet are also significantly better then training from scratch (up to 2\%, see Appendix~\ref{app:broader_augmentation}).

\textbf{Transfer learning.} We evaluate transfer learning performance across 12 natural image datasets in both linear evaluation (fixed feature extractor) and fine-tuning settings. Following \citet{kornblith2019better}, we perform hyperparameter tuning for each model-dataset combination and select the best hyperparameters on a validation set. Table~\ref{tab:transfer_learning_resnet_4x} shows results with the ResNet-50 ($4\times$) model.  When fine-tuned, our self-supervised model significantly outperforms the supervised baseline on 5 datasets, whereas the supervised baseline is superior on only 2 (i.e. Pets and Flowers). On the remaining 5 datasets, the models are statistically tied. 
Full experimental details as well as results with the standard ResNet-50 architecture are provided in Appendix~\ref{app:transfer_learning}. \section{Related Work}
\label{sec:related}

The idea of making representations of an image agree with each other under small transformations dates back to \citet{becker1992self}. We extend it by leveraging recent advances in data augmentation, network architecture and contrastive loss. A similar consistency idea, but for \textit{class label prediction}, has been explored in other contexts such as semi-supervised learning~\cite{xie2019unsupervised,berthelot2019mixmatch}.

\textbf{Handcrafted pretext tasks.} The recent renaissance of self-supervised learning began with artificially designed pretext tasks, such as relative patch prediction~\cite{doersch2015unsupervised}, solving jigsaw puzzles~\cite{noroozi2016unsupervised}, colorization~\cite{zhang2016colorful} and rotation prediction~\cite{gidaris2018unsupervised,chen2019self}. Although good results can be obtained with bigger networks and longer training~\cite{kolesnikov2019revisiting}, these pretext tasks rely on somewhat ad-hoc heuristics, which limits the generality of learned representations.

\textbf{Contrastive visual representation learning.} Dating back to~\citet{hadsell2006dimensionality}, these approaches learn representations by contrasting positive pairs against negative pairs. Along these lines, \citet{dosovitskiy2014discriminative} proposes to treat each instance as a class represented by a feature vector (in a parametric form). \citet{wu2018unsupervised} proposes to use a memory bank to store the instance class representation vector, an approach adopted and extended in several recent papers ~\cite{zhuang2019local,tian2019contrastive,he2019momentum,misra2019self}. Other work explores the use of in-batch samples for negative sampling instead of a memory bank~\cite{doersch2017multi,ye2019unsupervised,ji2019invariant}.

Recent literature has attempted to relate the success of their methods to maximization of mutual information between latent representations~\cite{oord2018representation,henaff2019data,hjelm2018learning,bachman2019learning}. However, it is not clear if the success of contrastive approaches is determined by the mutual information, or by the specific form of the contrastive loss~\cite{tschannen2019mutual}. 

We note that almost all individual components of our framework have appeared in previous work, although the specific instantiations may be different. The superiority of our framework relative to previous work is not explained by any single design choice, but by their composition. We provide a comprehensive comparison of our design choices with those of previous work in Appendix~\ref{app:related}.
 \vspace{-.5em}\section{Conclusion}
In this work, we present a simple framework and its instantiation for contrastive visual representation learning. We carefully study its components, and show the effects of different design choices. By combining our findings, we improve considerably over previous methods for self-supervised, semi-supervised, and transfer learning.

Our approach differs from standard supervised learning on ImageNet only in the choice of data augmentation, the use of a nonlinear head at the end of the network, and the loss function. The strength of this simple framework suggests that, despite a recent surge in interest, self-supervised learning remains undervalued. 
 
\section*{Acknowledgements}
We would like to thank Xiaohua Zhai, Rafael Müller and Yani Ioannou for their feedback on the draft.
We are also grateful for general support from Google Research teams in Toronto and elsewhere.

\FloatBarrier
{\small
\bibliography{content/ref}

\begin{thebibliography}{61}
\providecommand{\natexlab}[1]{#1}
\providecommand{\url}[1]{\texttt{#1}}
\expandafter\ifx\csname urlstyle\endcsname\relax
  \providecommand{\doi}[1]{doi: #1}\else
  \providecommand{\doi}{doi: \begingroup \urlstyle{rm}\Url}\fi

\bibitem[Asano et~al.(2019)Asano, Rupprecht, and Vedaldi]{asano2019critical}
Asano, Y.~M., Rupprecht, C., and Vedaldi, A.
\newblock A critical analysis of self-supervision, or what we can learn from a
  single image.
\newblock \emph{arXiv preprint arXiv:1904.13132}, 2019.

\bibitem[Bachman et~al.(2019)Bachman, Hjelm, and
  Buchwalter]{bachman2019learning}
Bachman, P., Hjelm, R.~D., and Buchwalter, W.
\newblock Learning representations by maximizing mutual information across
  views.
\newblock In \emph{Advances in Neural Information Processing Systems}, pp.\
  15509--15519, 2019.

\bibitem[Becker \& Hinton(1992)Becker and Hinton]{becker1992self}
Becker, S. and Hinton, G.~E.
\newblock Self-organizing neural network that discovers surfaces in random-dot
  stereograms.
\newblock \emph{Nature}, 355\penalty0 (6356):\penalty0 161--163, 1992.

\bibitem[Berg et~al.(2014)Berg, Liu, Lee, Alexander, Jacobs, and
  Belhumeur]{berg2014birdsnap}
Berg, T., Liu, J., Lee, S.~W., Alexander, M.~L., Jacobs, D.~W., and Belhumeur,
  P.~N.
\newblock Birdsnap: Large-scale fine-grained visual categorization of birds.
\newblock In \emph{IEEE Conference on Computer Vision and Pattern Recognition
  (CVPR)}, pp.\  2019--2026. IEEE, 2014.

\bibitem[Berthelot et~al.(2019)Berthelot, Carlini, Goodfellow, Papernot,
  Oliver, and Raffel]{berthelot2019mixmatch}
Berthelot, D., Carlini, N., Goodfellow, I., Papernot, N., Oliver, A., and
  Raffel, C.~A.
\newblock Mixmatch: A holistic approach to semi-supervised learning.
\newblock In \emph{Advances in Neural Information Processing Systems}, pp.\
  5050--5060, 2019.

\bibitem[Bossard et~al.(2014)Bossard, Guillaumin, and
  Van~Gool]{bossard2014food}
Bossard, L., Guillaumin, M., and Van~Gool, L.
\newblock Food-101--mining discriminative components with random forests.
\newblock In \emph{European conference on computer vision}, pp.\  446--461.
  Springer, 2014.

\bibitem[Chen et~al.(2017)Chen, Sun, Shi, and Hong]{chen2017sampling}
Chen, T., Sun, Y., Shi, Y., and Hong, L.
\newblock On sampling strategies for neural network-based collaborative
  filtering.
\newblock In \emph{Proceedings of the 23rd ACM SIGKDD International Conference
  on Knowledge Discovery and Data Mining}, pp.\  767--776, 2017.

\bibitem[Chen et~al.(2019)Chen, Zhai, Ritter, Lucic, and Houlsby]{chen2019self}
Chen, T., Zhai, X., Ritter, M., Lucic, M., and Houlsby, N.
\newblock Self-supervised gans via auxiliary rotation loss.
\newblock In \emph{Proceedings of the IEEE Conference on Computer Vision and
  Pattern Recognition}, pp.\  12154--12163, 2019.

\bibitem[Cimpoi et~al.(2014)Cimpoi, Maji, Kokkinos, Mohamed, and
  Vedaldi]{cimpoi2014describing}
Cimpoi, M., Maji, S., Kokkinos, I., Mohamed, S., and Vedaldi, A.
\newblock Describing textures in the wild.
\newblock In \emph{IEEE Conference on Computer Vision and Pattern Recognition
  (CVPR)}, pp.\  3606--3613. IEEE, 2014.

\bibitem[Cubuk et~al.(2019)Cubuk, Zoph, Mane, Vasudevan, and
  Le]{cubuk2019autoaugment}
Cubuk, E.~D., Zoph, B., Mane, D., Vasudevan, V., and Le, Q.~V.
\newblock Autoaugment: Learning augmentation strategies from data.
\newblock In \emph{Proceedings of the IEEE conference on computer vision and
  pattern recognition}, pp.\  113--123, 2019.

\bibitem[DeVries \& Taylor(2017)DeVries and Taylor]{devries2017improved}
DeVries, T. and Taylor, G.~W.
\newblock Improved regularization of convolutional neural networks with cutout.
\newblock \emph{arXiv preprint arXiv:1708.04552}, 2017.

\bibitem[Doersch \& Zisserman(2017)Doersch and Zisserman]{doersch2017multi}
Doersch, C. and Zisserman, A.
\newblock Multi-task self-supervised visual learning.
\newblock In \emph{Proceedings of the IEEE International Conference on Computer
  Vision}, pp.\  2051--2060, 2017.

\bibitem[Doersch et~al.(2015)Doersch, Gupta, and
  Efros]{doersch2015unsupervised}
Doersch, C., Gupta, A., and Efros, A.~A.
\newblock Unsupervised visual representation learning by context prediction.
\newblock In \emph{Proceedings of the IEEE International Conference on Computer
  Vision}, pp.\  1422--1430, 2015.

\bibitem[Donahue \& Simonyan(2019)Donahue and Simonyan]{donahue2019large}
Donahue, J. and Simonyan, K.
\newblock Large scale adversarial representation learning.
\newblock In \emph{Advances in Neural Information Processing Systems}, pp.\
  10541--10551, 2019.

\bibitem[Donahue et~al.(2014)Donahue, Jia, Vinyals, Hoffman, Zhang, Tzeng, and
  Darrell]{donahue2014decaf}
Donahue, J., Jia, Y., Vinyals, O., Hoffman, J., Zhang, N., Tzeng, E., and
  Darrell, T.
\newblock Decaf: A deep convolutional activation feature for generic visual
  recognition.
\newblock In \emph{International Conference on Machine Learning}, pp.\
  647--655, 2014.

\bibitem[Dosovitskiy et~al.(2014)Dosovitskiy, Springenberg, Riedmiller, and
  Brox]{dosovitskiy2014discriminative}
Dosovitskiy, A., Springenberg, J.~T., Riedmiller, M., and Brox, T.
\newblock Discriminative unsupervised feature learning with convolutional
  neural networks.
\newblock In \emph{Advances in neural information processing systems}, pp.\
  766--774, 2014.

\bibitem[Everingham et~al.(2010)Everingham, Van~Gool, Williams, Winn, and
  Zisserman]{everingham2010pascal}
Everingham, M., Van~Gool, L., Williams, C.~K., Winn, J., and Zisserman, A.
\newblock The pascal visual object classes (voc) challenge.
\newblock \emph{International Journal of Computer Vision}, 88\penalty0
  (2):\penalty0 303--338, 2010.

\bibitem[Fei-Fei et~al.(2004)Fei-Fei, Fergus, and Perona]{fei2004learning}
Fei-Fei, L., Fergus, R., and Perona, P.
\newblock Learning generative visual models from few training examples: An
  incremental bayesian approach tested on 101 object categories.
\newblock In \emph{IEEE Conference on Computer Vision and Pattern Recognition
  (CVPR) Workshop on Generative-Model Based Vision}, 2004.

\bibitem[Gidaris et~al.(2018)Gidaris, Singh, and
  Komodakis]{gidaris2018unsupervised}
Gidaris, S., Singh, P., and Komodakis, N.
\newblock Unsupervised representation learning by predicting image rotations.
\newblock \emph{arXiv preprint arXiv:1803.07728}, 2018.

\bibitem[Goodfellow et~al.(2014)Goodfellow, Pouget-Abadie, Mirza, Xu,
  Warde-Farley, Ozair, Courville, and Bengio]{goodfellow2014generative}
Goodfellow, I., Pouget-Abadie, J., Mirza, M., Xu, B., Warde-Farley, D., Ozair,
  S., Courville, A., and Bengio, Y.
\newblock Generative adversarial nets.
\newblock In \emph{Advances in neural information processing systems}, pp.\
  2672--2680, 2014.

\bibitem[Goyal et~al.(2017)Goyal, Doll{\'a}r, Girshick, Noordhuis, Wesolowski,
  Kyrola, Tulloch, Jia, and He]{goyal2017accurate}
Goyal, P., Doll{\'a}r, P., Girshick, R., Noordhuis, P., Wesolowski, L., Kyrola,
  A., Tulloch, A., Jia, Y., and He, K.
\newblock Accurate, large minibatch sgd: Training imagenet in 1 hour.
\newblock \emph{arXiv preprint arXiv:1706.02677}, 2017.

\bibitem[Hadsell et~al.(2006)Hadsell, Chopra, and
  LeCun]{hadsell2006dimensionality}
Hadsell, R., Chopra, S., and LeCun, Y.
\newblock Dimensionality reduction by learning an invariant mapping.
\newblock In \emph{2006 IEEE Computer Society Conference on Computer Vision and
  Pattern Recognition (CVPR'06)}, volume~2, pp.\  1735--1742. IEEE, 2006.

\bibitem[He et~al.(2016)He, Zhang, Ren, and Sun]{he2016deep}
He, K., Zhang, X., Ren, S., and Sun, J.
\newblock Deep residual learning for image recognition.
\newblock In \emph{Proceedings of the IEEE conference on computer vision and
  pattern recognition}, pp.\  770--778, 2016.

\bibitem[He et~al.(2019)He, Fan, Wu, Xie, and Girshick]{he2019momentum}
He, K., Fan, H., Wu, Y., Xie, S., and Girshick, R.
\newblock Momentum contrast for unsupervised visual representation learning.
\newblock \emph{arXiv preprint arXiv:1911.05722}, 2019.

\bibitem[H{\'e}naff et~al.(2019)H{\'e}naff, Razavi, Doersch, Eslami, and
  Oord]{henaff2019data}
H{\'e}naff, O.~J., Razavi, A., Doersch, C., Eslami, S., and Oord, A. v.~d.
\newblock Data-efficient image recognition with contrastive predictive coding.
\newblock \emph{arXiv preprint arXiv:1905.09272}, 2019.

\bibitem[Hinton et~al.(2006)Hinton, Osindero, and Teh]{hinton2006fast}
Hinton, G.~E., Osindero, S., and Teh, Y.-W.
\newblock A fast learning algorithm for deep belief nets.
\newblock \emph{Neural computation}, 18\penalty0 (7):\penalty0 1527--1554,
  2006.

\bibitem[Hjelm et~al.(2018)Hjelm, Fedorov, Lavoie-Marchildon, Grewal, Bachman,
  Trischler, and Bengio]{hjelm2018learning}
Hjelm, R.~D., Fedorov, A., Lavoie-Marchildon, S., Grewal, K., Bachman, P.,
  Trischler, A., and Bengio, Y.
\newblock Learning deep representations by mutual information estimation and
  maximization.
\newblock \emph{arXiv preprint arXiv:1808.06670}, 2018.

\bibitem[Howard(2013)]{howard2013some}
Howard, A.~G.
\newblock Some improvements on deep convolutional neural network based image
  classification.
\newblock \emph{arXiv preprint arXiv:1312.5402}, 2013.

\bibitem[Ioffe \& Szegedy(2015)Ioffe and Szegedy]{ioffe2015batch}
Ioffe, S. and Szegedy, C.
\newblock Batch normalization: Accelerating deep network training by reducing
  internal covariate shift.
\newblock \emph{arXiv preprint arXiv:1502.03167}, 2015.

\bibitem[Ji et~al.(2019)Ji, Henriques, and Vedaldi]{ji2019invariant}
Ji, X., Henriques, J.~F., and Vedaldi, A.
\newblock Invariant information clustering for unsupervised image
  classification and segmentation.
\newblock In \emph{Proceedings of the IEEE International Conference on Computer
  Vision}, pp.\  9865--9874, 2019.

\bibitem[Kingma \& Welling(2013)Kingma and Welling]{kingma2013auto}
Kingma, D.~P. and Welling, M.
\newblock Auto-encoding variational bayes.
\newblock \emph{arXiv preprint arXiv:1312.6114}, 2013.

\bibitem[Kolesnikov et~al.(2019)Kolesnikov, Zhai, and
  Beyer]{kolesnikov2019revisiting}
Kolesnikov, A., Zhai, X., and Beyer, L.
\newblock Revisiting self-supervised visual representation learning.
\newblock In \emph{Proceedings of the IEEE conference on Computer Vision and
  Pattern Recognition}, pp.\  1920--1929, 2019.

\bibitem[Kornblith et~al.(2019)Kornblith, Shlens, and Le]{kornblith2019better}
Kornblith, S., Shlens, J., and Le, Q.~V.
\newblock Do better {ImageNet} models transfer better?
\newblock In \emph{Proceedings of the IEEE conference on computer vision and
  pattern recognition}, pp.\  2661--2671, 2019.

\bibitem[Krause et~al.(2013)Krause, Deng, Stark, and
  Fei-Fei]{krause2013collecting}
Krause, J., Deng, J., Stark, M., and Fei-Fei, L.
\newblock Collecting a large-scale dataset of fine-grained cars.
\newblock In \emph{Second Workshop on Fine-Grained Visual Categorization},
  2013.

\bibitem[Krizhevsky \& Hinton(2009)Krizhevsky and
  Hinton]{krizhevsky2009learning}
Krizhevsky, A. and Hinton, G.
\newblock Learning multiple layers of features from tiny images.
\newblock Technical report, University of Toronto, 2009.
\newblock URL
  \url{https://www.cs.toronto.edu/~kriz/learning-features-2009-TR.pdf}.

\bibitem[Krizhevsky et~al.(2012)Krizhevsky, Sutskever, and
  Hinton]{krizhevsky2012imagenet}
Krizhevsky, A., Sutskever, I., and Hinton, G.~E.
\newblock Imagenet classification with deep convolutional neural networks.
\newblock In \emph{Advances in neural information processing systems}, pp.\
  1097--1105, 2012.

\bibitem[Loshchilov \& Hutter(2016)Loshchilov and Hutter]{loshchilov2016sgdr}
Loshchilov, I. and Hutter, F.
\newblock Sgdr: Stochastic gradient descent with warm restarts.
\newblock \emph{arXiv preprint arXiv:1608.03983}, 2016.

\bibitem[Maaten \& Hinton(2008)Maaten and Hinton]{maaten2008visualizing}
Maaten, L. v.~d. and Hinton, G.
\newblock Visualizing data using t-sne.
\newblock \emph{Journal of machine learning research}, 9\penalty0
  (Nov):\penalty0 2579--2605, 2008.

\bibitem[Maji et~al.(2013)Maji, Kannala, Rahtu, Blaschko, and
  Vedaldi]{maji13fine-grained}
Maji, S., Kannala, J., Rahtu, E., Blaschko, M., and Vedaldi, A.
\newblock Fine-grained visual classification of aircraft.
\newblock Technical report, 2013.

\bibitem[Mikolov et~al.(2013)Mikolov, Chen, Corrado, and
  Dean]{mikolov2013efficient}
Mikolov, T., Chen, K., Corrado, G., and Dean, J.
\newblock Efficient estimation of word representations in vector space.
\newblock \emph{arXiv preprint arXiv:1301.3781}, 2013.

\bibitem[Misra \& van~der Maaten(2019)Misra and van~der Maaten]{misra2019self}
Misra, I. and van~der Maaten, L.
\newblock Self-supervised learning of pretext-invariant representations.
\newblock \emph{arXiv preprint arXiv:1912.01991}, 2019.

\bibitem[Nilsback \& Zisserman(2008)Nilsback and
  Zisserman]{nilsback2008automated}
Nilsback, M.-E. and Zisserman, A.
\newblock Automated flower classification over a large number of classes.
\newblock In \emph{Computer Vision, Graphics \& Image Processing, 2008.
  ICVGIP'08. Sixth Indian Conference on}, pp.\  722--729. IEEE, 2008.

\bibitem[Noroozi \& Favaro(2016)Noroozi and Favaro]{noroozi2016unsupervised}
Noroozi, M. and Favaro, P.
\newblock Unsupervised learning of visual representations by solving jigsaw
  puzzles.
\newblock In \emph{European Conference on Computer Vision}, pp.\  69--84.
  Springer, 2016.

\bibitem[Oord et~al.(2018)Oord, Li, and Vinyals]{oord2018representation}
Oord, A. v.~d., Li, Y., and Vinyals, O.
\newblock Representation learning with contrastive predictive coding.
\newblock \emph{arXiv preprint arXiv:1807.03748}, 2018.

\bibitem[Parkhi et~al.(2012)Parkhi, Vedaldi, Zisserman, and
  Jawahar]{parkhi2012cats}
Parkhi, O.~M., Vedaldi, A., Zisserman, A., and Jawahar, C.
\newblock Cats and dogs.
\newblock In \emph{IEEE Conference on Computer Vision and Pattern Recognition
  (CVPR)}, pp.\  3498--3505. IEEE, 2012.

\bibitem[Russakovsky et~al.(2015)Russakovsky, Deng, Su, Krause, Satheesh, Ma,
  Huang, Karpathy, Khosla, Bernstein, et~al.]{russakovsky2015imagenet}
Russakovsky, O., Deng, J., Su, H., Krause, J., Satheesh, S., Ma, S., Huang, Z.,
  Karpathy, A., Khosla, A., Bernstein, M., et~al.
\newblock Imagenet large scale visual recognition challenge.
\newblock \emph{International journal of computer vision}, 115\penalty0
  (3):\penalty0 211--252, 2015.

\bibitem[Schroff et~al.(2015)Schroff, Kalenichenko, and
  Philbin]{schroff2015facenet}
Schroff, F., Kalenichenko, D., and Philbin, J.
\newblock Facenet: A unified embedding for face recognition and clustering.
\newblock In \emph{Proceedings of the IEEE conference on computer vision and
  pattern recognition}, pp.\  815--823, 2015.

\bibitem[Simonyan \& Zisserman(2014)Simonyan and Zisserman]{simonyan2014very}
Simonyan, K. and Zisserman, A.
\newblock Very deep convolutional networks for large-scale image recognition.
\newblock \emph{arXiv preprint arXiv:1409.1556}, 2014.

\bibitem[Sohn(2016)]{sohn2016improved}
Sohn, K.
\newblock Improved deep metric learning with multi-class n-pair loss objective.
\newblock In \emph{Advances in neural information processing systems}, pp.\
  1857--1865, 2016.

\bibitem[Sohn et~al.(2020)Sohn, Berthelot, Li, Zhang, Carlini, Cubuk, Kurakin,
  Zhang, and Raffel]{sohn2020fixmatch}
Sohn, K., Berthelot, D., Li, C.-L., Zhang, Z., Carlini, N., Cubuk, E.~D.,
  Kurakin, A., Zhang, H., and Raffel, C.
\newblock Fixmatch: Simplifying semi-supervised learning with consistency and
  confidence.
\newblock \emph{arXiv preprint arXiv:2001.07685}, 2020.

\bibitem[Szegedy et~al.(2015)Szegedy, Liu, Jia, Sermanet, Reed, Anguelov,
  Erhan, Vanhoucke, and Rabinovich]{szegedy2015going}
Szegedy, C., Liu, W., Jia, Y., Sermanet, P., Reed, S., Anguelov, D., Erhan, D.,
  Vanhoucke, V., and Rabinovich, A.
\newblock Going deeper with convolutions.
\newblock In \emph{Proceedings of the IEEE conference on computer vision and
  pattern recognition}, pp.\  1--9, 2015.

\bibitem[Tian et~al.(2019)Tian, Krishnan, and Isola]{tian2019contrastive}
Tian, Y., Krishnan, D., and Isola, P.
\newblock Contrastive multiview coding.
\newblock \emph{arXiv preprint arXiv:1906.05849}, 2019.

\bibitem[Tschannen et~al.(2019)Tschannen, Djolonga, Rubenstein, Gelly, and
  Lucic]{tschannen2019mutual}
Tschannen, M., Djolonga, J., Rubenstein, P.~K., Gelly, S., and Lucic, M.
\newblock On mutual information maximization for representation learning.
\newblock \emph{arXiv preprint arXiv:1907.13625}, 2019.

\bibitem[Wu et~al.(2018)Wu, Xiong, Yu, and Lin]{wu2018unsupervised}
Wu, Z., Xiong, Y., Yu, S.~X., and Lin, D.
\newblock Unsupervised feature learning via non-parametric instance
  discrimination.
\newblock In \emph{Proceedings of the IEEE Conference on Computer Vision and
  Pattern Recognition}, pp.\  3733--3742, 2018.

\bibitem[Xiao et~al.(2010)Xiao, Hays, Ehinger, Oliva, and
  Torralba]{xiao2010sun}
Xiao, J., Hays, J., Ehinger, K.~A., Oliva, A., and Torralba, A.
\newblock Sun database: Large-scale scene recognition from abbey to zoo.
\newblock In \emph{IEEE Conference on Computer Vision and Pattern Recognition
  (CVPR)}, pp.\  3485--3492. IEEE, 2010.

\bibitem[Xie et~al.(2019)Xie, Dai, Hovy, Luong, and Le]{xie2019unsupervised}
Xie, Q., Dai, Z., Hovy, E., Luong, M.-T., and Le, Q.~V.
\newblock Unsupervised data augmentation.
\newblock \emph{arXiv preprint arXiv:1904.12848}, 2019.

\bibitem[Ye et~al.(2019)Ye, Zhang, Yuen, and Chang]{ye2019unsupervised}
Ye, M., Zhang, X., Yuen, P.~C., and Chang, S.-F.
\newblock Unsupervised embedding learning via invariant and spreading instance
  feature.
\newblock In \emph{Proceedings of the IEEE Conference on Computer Vision and
  Pattern Recognition}, pp.\  6210--6219, 2019.

\bibitem[You et~al.(2017)You, Gitman, and Ginsburg]{you2017large}
You, Y., Gitman, I., and Ginsburg, B.
\newblock Large batch training of convolutional networks.
\newblock \emph{arXiv preprint arXiv:1708.03888}, 2017.

\bibitem[Zhai et~al.(2019)Zhai, Oliver, Kolesnikov, and Beyer]{zhai2019s4l}
Zhai, X., Oliver, A., Kolesnikov, A., and Beyer, L.
\newblock S4l: Self-supervised semi-supervised learning.
\newblock In \emph{The IEEE International Conference on Computer Vision
  (ICCV)}, October 2019.

\bibitem[Zhang et~al.(2016)Zhang, Isola, and Efros]{zhang2016colorful}
Zhang, R., Isola, P., and Efros, A.~A.
\newblock Colorful image colorization.
\newblock In \emph{European conference on computer vision}, pp.\  649--666.
  Springer, 2016.

\bibitem[Zhuang et~al.(2019)Zhuang, Zhai, and Yamins]{zhuang2019local}
Zhuang, C., Zhai, A.~L., and Yamins, D.
\newblock Local aggregation for unsupervised learning of visual embeddings.
\newblock In \emph{Proceedings of the IEEE International Conference on Computer
  Vision}, pp.\  6002--6012, 2019.

\end{thebibliography}
\bibliographystyle{icml2020}}

\cleardoublepage
\appendix
\counterwithin{figure}{section}
\counterwithin{table}{section}
\onecolumn

\appendix
\section{Data Augmentation Details}
\label{app:da}

In our default pretraining setting (which is used to train our best models), we utilize random crop (with resize and random flip), random color distortion, and random Gaussian blur as the data augmentations. The details of these three augmentations are provided below.

\paragraph{Random crop and resize to 224x224} We use standard Inception-style random cropping~\cite{szegedy2015going}. The crop of random size (uniform from 0.08 to 1.0 in area) of the original size and a random aspect ratio (default: of 3/4 to 4/3) of the original aspect ratio is made. This crop is finally resized to the original size. This has been implemented in Tensorflow as ``$\mathrm{slim.preprocessing.inception\_preprocessing.distorted\_bounding\_box\_crop}$'', or in Pytorch as ``$\mathrm{torchvision.transforms.RandomResizedCrop}$''. Additionally, the random crop (with resize) is always followed by a random horizontal/left-to-right flip with $50\%$ probability. This is helpful but not essential. By removing this from our default augmentation policy, the top-1 linear evaluation drops from 64.5\% to 63.4\% for our ResNet-50 model trained in 100 epochs.

\paragraph{Color distortion} Color distortion is composed by color jittering and color dropping. We find stronger color jittering usually helps, so we set a strength parameter. 

A pseudo-code for color distortion using TensorFlow is as follows.
\begin{verbatim}
import tensorflow as tf
def color_distortion(image, s=1.0):
    # image is a tensor with value range in [0, 1].
    # s is the strength of color distortion.
    
    def color_jitter(x):
        # one can also shuffle the order of following augmentations
        # each time they are applied.
        x = tf.image.random_brightness(x, max_delta=0.8*s)
        x = tf.image.random_contrast(x, lower=1-0.8*s, upper=1+0.8*s)
        x = tf.image.random_saturation(x, lower=1-0.8*s, upper=1+0.8*s)
        x = tf.image.random_hue(x, max_delta=0.2*s)
        x = tf.clip_by_value(x, 0, 1)
        return x
    
    def color_drop(x):
        image = tf.image.rgb_to_grayscale(image)
        image = tf.tile(image, [1, 1, 3])
        
    # randomly apply transformation with probability p.
    image = random_apply(color_jitter, image, p=0.8)
    image = random_apply(color_drop, image, p=0.2)
    return image
\end{verbatim}

A pseudo-code for color distortion using Pytorch is as follows~\footnote{Our code and results are based on Tensorflow, the Pytorch code here is a reference.}.
\begin{verbatim}
from torchvision import transforms
def get_color_distortion(s=1.0):
    # s is the strength of color distortion.
    color_jitter = transforms.ColorJitter(0.8*s, 0.8*s, 0.8*s, 0.2*s)
    rnd_color_jitter = transforms.RandomApply([color_jitter], p=0.8)
    rnd_gray = transforms.RandomGrayscale(p=0.2)
    color_distort = transforms.Compose([
        rnd_color_jitter,
        rnd_gray])
    return color_distort
\end{verbatim}

\paragraph{Gaussian blur}

This augmentation is in our default policy. We find it helpful, as it improves our ResNet-50 trained for 100 epochs from 63.2\% to 64.5\%. We blur the image 50\% of the time using a Gaussian kernel. We randomly sample $\sigma\in[0.1, 2.0]$, and the kernel size is set to be 10\% of the image height/width.

\section{Additional Experimental Results}
\label{app:result}

\subsection{Batch Size and Training Steps}
\label{app:bsz_step}

\begin{figure}[h!]
    \begin{minipage}[c]{0.48\linewidth}
        \centering
        \includegraphics[width=1.0\linewidth]{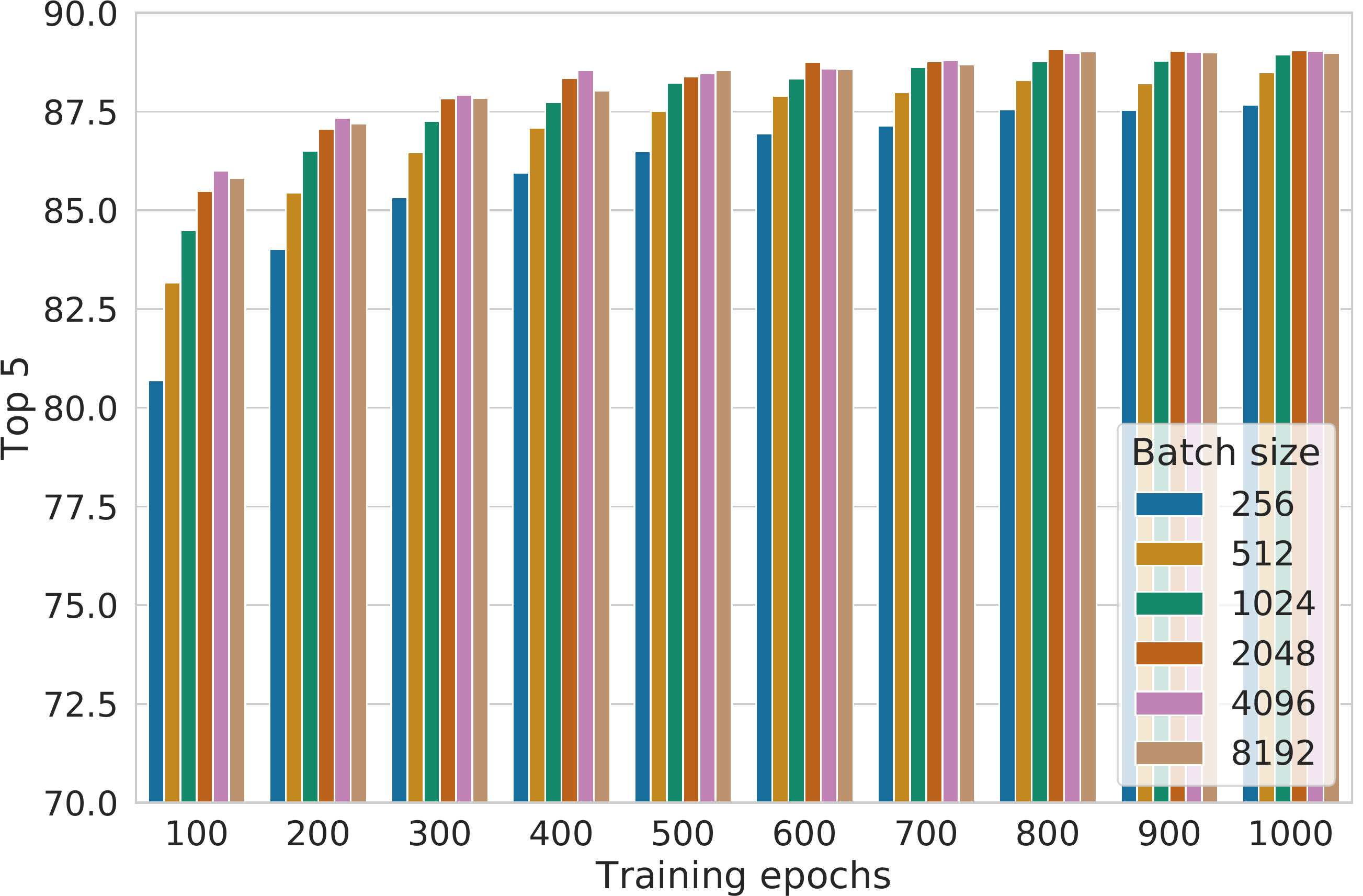}
        \caption{\label{fig:bsstep_top5}Linear evaluation (top-5) of ResNet-50 trained with different batch sizes and epochs. Each bar is a single run from scratch. See Figure~\ref{fig:bsstep_top1} for top-1 accuracy.}
    \end{minipage}\hfill
    \begin{minipage}[c]{0.48\linewidth}
    \centering
    \includegraphics[width=1.0\textwidth]{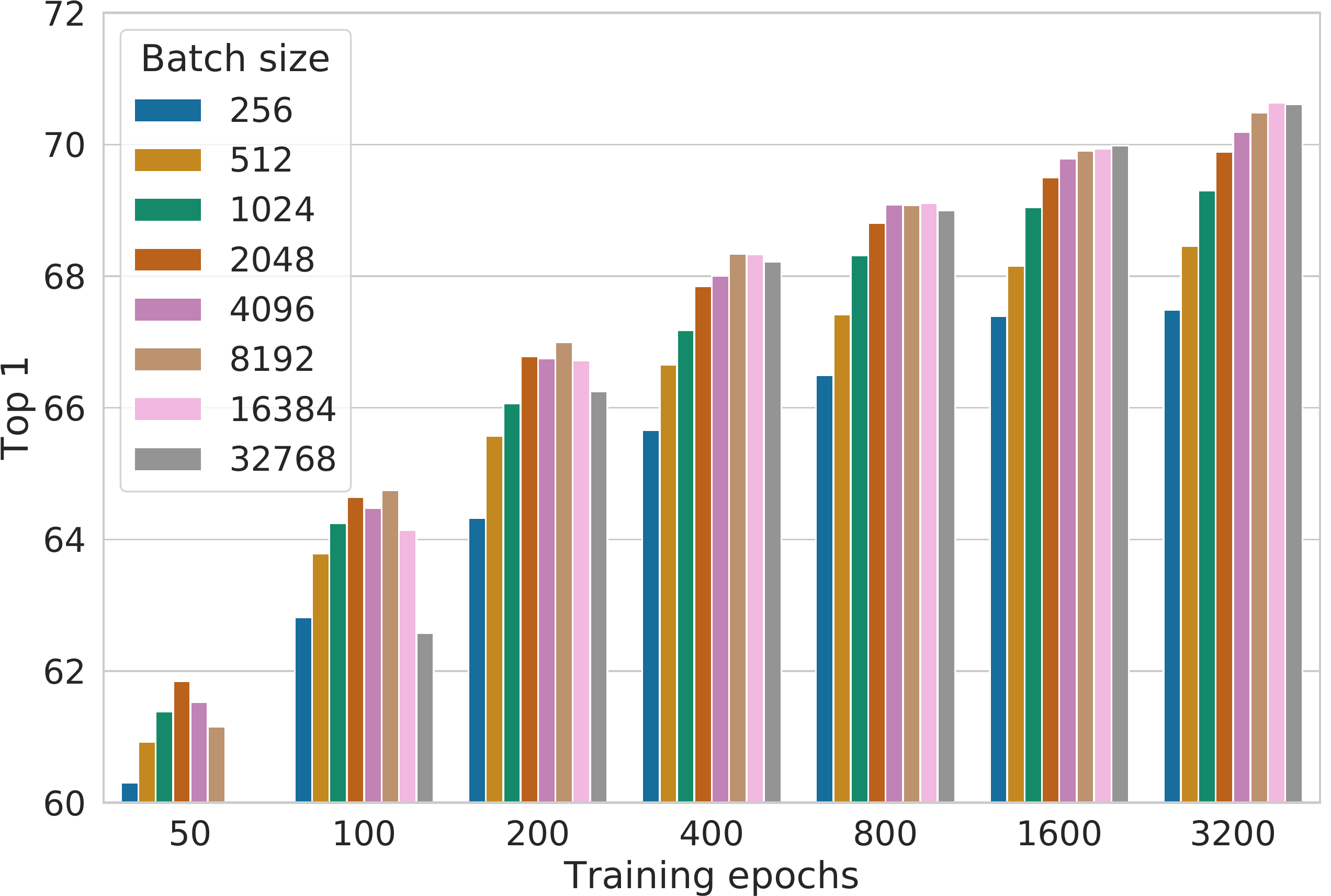}
    \caption{Linear evaluation (top-1) of ResNet-50 trained with different batch sizes and \textit{longer} epochs. Here a \textit{square root} learning rate, instead of a linear one, is utilized.}
    \label{fig:bsstep_top1_longer}
    \end{minipage}
\end{figure}

Figure~\ref{fig:bsstep_top5} shows the top-5 accuracy on linear evaluation when trained with different batch sizes and training epochs. The conclusion is very similar to top-1 accuracy shown before, except that the differences between different batch sizes and training steps seems slightly smaller here.

In both Figure~\ref{fig:bsstep_top1} and Figure~\ref{fig:bsstep_top5}, we use a linear scaling of learning rate similar to~\cite{goyal2017accurate} when training with different batch sizes. Although linear learning rate scaling is popular with SGD/Momentum optimizer, we find a square root learning rate scaling is more desirable with LARS optimizer. With square root learning rate scaling, we have $\mathrm{LearningRate}=0.075\times\sqrt\mathrm{BatchSize}$, instead of $\mathrm{LearningRate}=0.3\times\mathrm{BatchSize}/256$ in the linear scaling case, but the learning rate is the same under both scaling methods when batch size of 4096 (our default batch size). A comparison is presented in Table~\ref{tab:sqrt_vs_linear_scaling}, where we observe that square root learning rate scaling improves the performance for models trained with small batch sizes and in smaller number of epochs.

\begin{table}[h!]
    \centering
    \small
    \begin{tabular}{c|ccccc}
    \toprule
    Batch size \textbackslash~ Epochs & 100 & 200 & 400 & 800 \\ \midrule
    256 & 57.5 / \textbf{62.8} & 61.9 / \textbf{64.3} & 64.7 / \textbf{65.7} & 66.6 / 66.5 \\
    512 & 60.7 / \textbf{63.8} & 64.0 / \textbf{65.6} & 66.2 / 66.7 & 67.8 / 67.4 \\
    1024 & 62.8 / \textbf{64.3} & 65.3 / \textbf{66.1} & 67.2 / 67.2 & 68.5 / 68.3 \\
    2048 & 64.0 / \textbf{64.7} & 66.1 / \textbf{66.8} & 68.1 / 67.9 & 68.9 / 68.8 \\
    4096 & 64.6 / 64.5 & 66.5 / 66.8 & 68.2 / 68.0 & 68.9 / 69.1 \\
    8192 & 64.8 / 64.8 & 66.6 / 67.0 & 67.8 / 68.3 & 69.0 / 69.1 \\
    \bottomrule
    \end{tabular}
    \caption{Linear evaluation (top-1) under different batch sizes and training epochs. On the left side of slash sign are models trained with linear LR scaling, and on the right are models trained with square root LR scaling. The result is bolded if it is more than 0.5\% better. Square root LR scaling works better for smaller batch size trained in fewer epochs (with LARS optimizer).}
    \label{tab:sqrt_vs_linear_scaling}
\end{table}

We also train with larger batch size (up to 32K) and longer (up to 3200 epochs), with the square root learning rate scaling. A shown in Figure~\ref{fig:bsstep_top1_longer}, \textit{the performance seems to saturate with a batch size of 8192, while training longer can still significantly improve the performance}.

\subsection{Broader composition of data augmentations further improves performance}
\label{app:broader_augmentation}

Our best results in the main text (Table~\ref{tab:linear_sota} and \ref{tab:data_efficient_sota}) can be further improved when expanding the default augmentation policy to include the following: (1) Sobel filtering, (2) additional color distortion (equalize, solarize), and (3) motion blur. For linear evaluation protocol, the ResNet-50 models ($1\times, 2\times, 4\times$) trained with broader data augmentations achieve 70.0 (+0.7), 74.4 (+0.2), 76.8 (+0.3), respectively.

Table~\ref{tab:data_efficient_sota_broader} shows ImageNet accuracy obtained by fine-tuning the SimCLR model (see Appendix~\ref{app:result_semi} for the details of fine-tuning procedure). Interestingly, when fine-tuned on full (100\%) ImageNet training set, our ResNet (4$\times$) model achieves 80.4\% top-1 / 95.4\% top-5~\footnote{It is 80.1\% top-1 / 95.2\% top-5 without broader augmentations for pretraining SimCLR.}, which is significantly better than that (78.4\% top-1 / 94.2\% top-5) of training from scratch using the same set of augmentations (i.e. random crop and horizontal flip). For ResNet-50 (2$\times$), fine-tuning our pre-trained ResNet-50 (2$\times$) is also better than training from scratch (77.8\% top-1 / 93.9\% top-5). There is no improvement from fine-tuning for ResNet-50.

\begin{table}[!h]
    \centering
    \begin{tabular}{lcccccc}
    \toprule
    \multirow{3}{*}{Architecture} & \multicolumn{6}{c}{Label fraction} \\
    & \multicolumn{2}{c}{1\%} & \multicolumn{2}{c}{10\%} & \multicolumn{2}{c}{100\%}  \\
    & Top 1 & Top 5 & Top 1 & Top 5 & Top 1 & Top 5\\  \midrule
    ResNet-50 & 49.4 & 	76.6 & 	66.1 & 88.1 & 76.0 & 93.1\\
    ResNet-50 (2$\times$) & 59.4 & 83.7 & 71.8 & 91.2 & 79.1 & 94.8 \\
    ResNet-50 (4$\times$) & 64.1 & 86.6 & 74.8 & 92.8 & 80.4 & 95.4\\
    \bottomrule
    \end{tabular}
    \caption{Classification accuracy obtained by fine-tuning the SimCLR (which is pretrained with broader data augmentations) on 1\%, 10\% and full of ImageNet. As a reference, our ResNet-50 (4$\times$) trained from scratch on 100\% labels achieves 78.4\% top-1 / 94.2\% top-5.}
    \label{tab:data_efficient_sota_broader}
\end{table}

\subsection{Effects of Longer Training for Supervised Models}
\label{app:sup}

Here we perform experiments to see how training steps and stronger data augmentation affect supervised training. We test ResNet-50 and ResNet-50 (4$\times$) under the same set of data augmentations (random crops, color distortion, 50\% Gaussian blur) as used in our unsupervised models. Figure~\ref{tab:sup_long} shows the top-1 accuracy. We observe that there is no significant benefit from training supervised models longer on ImageNet. Stronger data augmentation slightly improves the accuracy of ResNet-50 (4$\times$) but does not help on ResNet-50. When stronger data augmentation is applied, ResNet-50 generally requires longer training (e.g. 500 epochs~\footnote{With AutoAugment~\cite{cubuk2019autoaugment}, optimal test accuracy can be achieved between 900 and 500 epochs.}) to obtain the optimal result, while ResNet-50 (4$\times$) does not benefit from longer  training. 

\begin{table}[h!]
\small
\centering
\begin{tabular}{ccccc}
\toprule
\multirow{2}{*}{Model} & \multirow{2}{*}{Training epochs} & \multicolumn{3}{c}{Top 1}     \\
                      &                         & Crop  & +Color & +Color+Blur \\\midrule 
\multirow{3}{*}{\rotatebox[origin=c]{0}{ResNet-50}}     & 90   & 76.5 & 75.6    & 75.3         \\ 
                          & 500        & 76.2 & 76.5    & 76.7         \\ 
                          & 1000        & 75.8 & 75.2    & 76.4         \\ \midrule
\multirow{3}{*}{\rotatebox[origin=c]{0}{ResNet-50 (4$\times$)}}& 90 &     78.4    & 78.9    & 78.7            \\
                          & 500 &     78.3    & 78.4    & 78.5            \\
                          & 1000  & 77.9 & 78.2    & 78.3       \\ \bottomrule
\end{tabular}
\caption{\label{tab:sup_long} Top-1 accuracy of supervised models trained longer under various data augmentation procedures (from the same set of data augmentations for contrastive learning).}
\end{table}

\subsection{Understanding The Non-Linear Projection Head}
\label{app:understand_head}

\begin{figure}[h!]
    \begin{minipage}[c]{0.49\linewidth}
        \centering
        \begin{subfigure}{.45\textwidth}
          \centering
          \includegraphics[width=\linewidth]{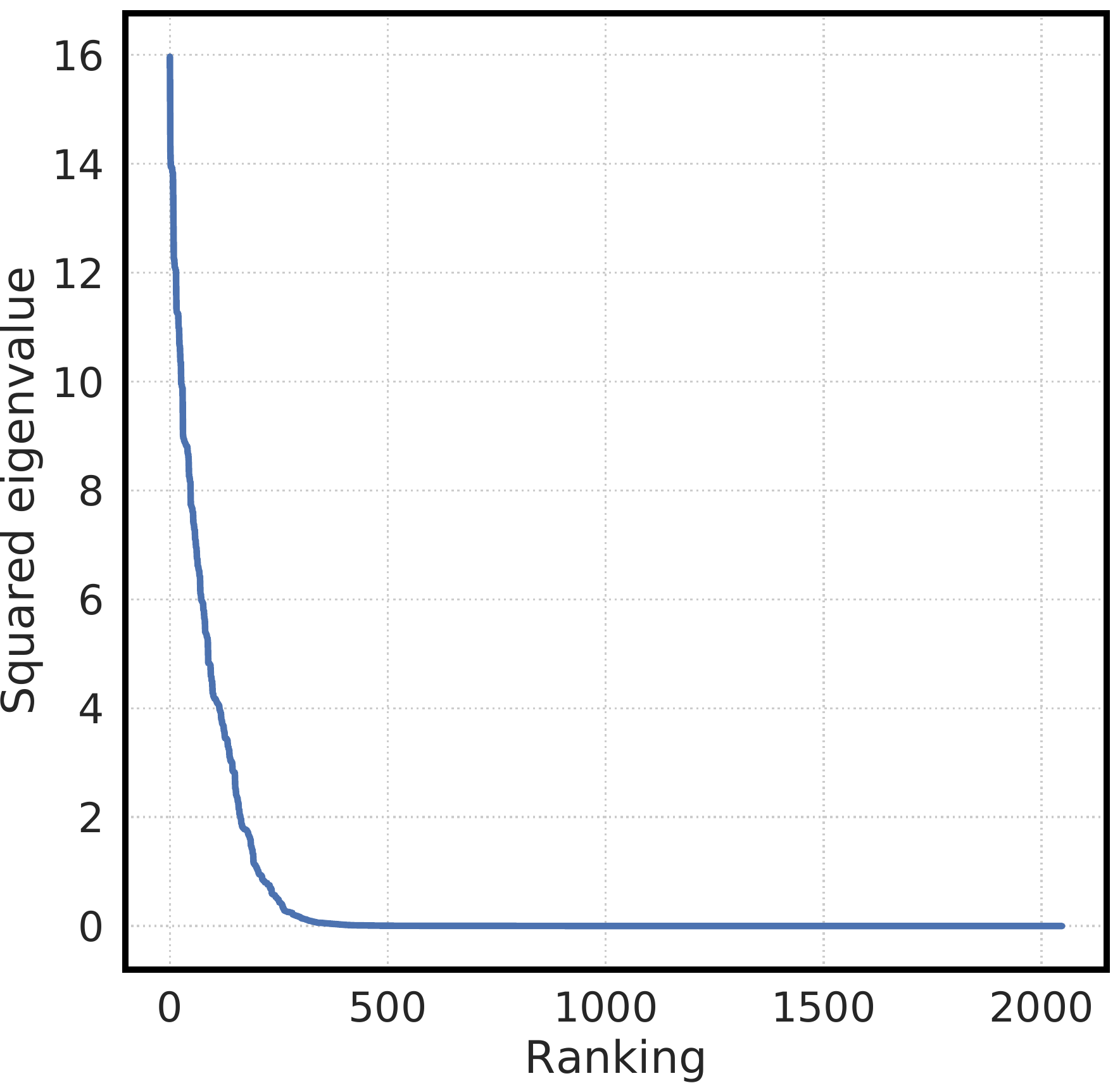}
          \caption{Y-axis in uniform scale.}
        \end{subfigure}~
        \begin{subfigure}{.48\textwidth}
          \centering
          \includegraphics[width=\linewidth]{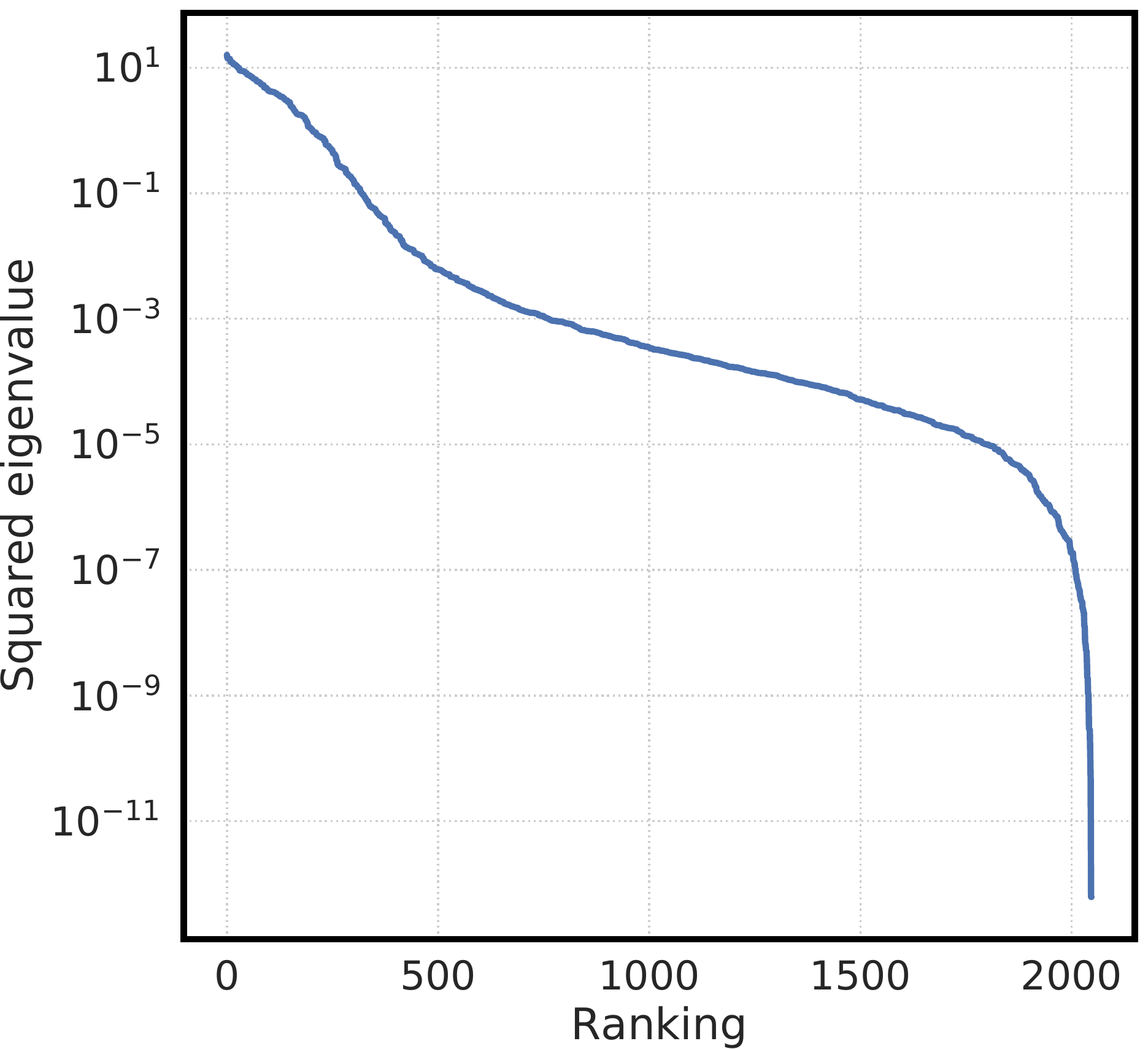}
          \caption{Y-axis in log scale.}
        \end{subfigure}\caption{\label{fig:critic_linproj_eig} Squared real eigenvalue distribution of linear projection matrix $W\in R^{2048\times2048}$ used to compute $g(\bm h) = W\bm h$.}
    \end{minipage}\hfill
    \begin{minipage}[c]{0.49\linewidth}
        \centering
        \begin{subfigure}{.48\textwidth}
          \centering
          \includegraphics[trim=50 0 50 50,clip,width=\linewidth]{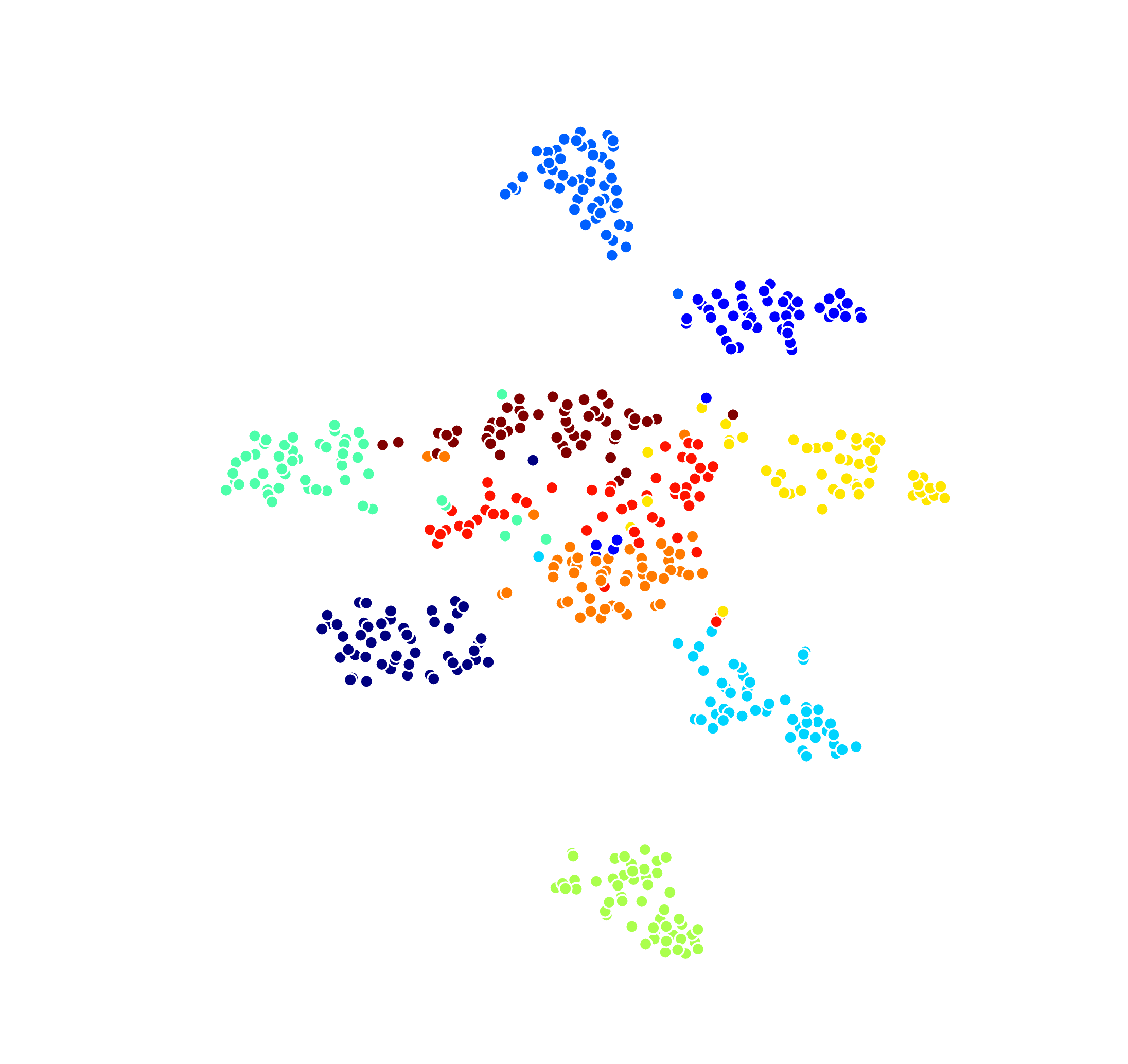}
          \caption{$\bm h$}
        \end{subfigure}~
        \begin{subfigure}{.48\textwidth}
          \centering
          \includegraphics[trim=30 0 30 20,clip,width=\linewidth]{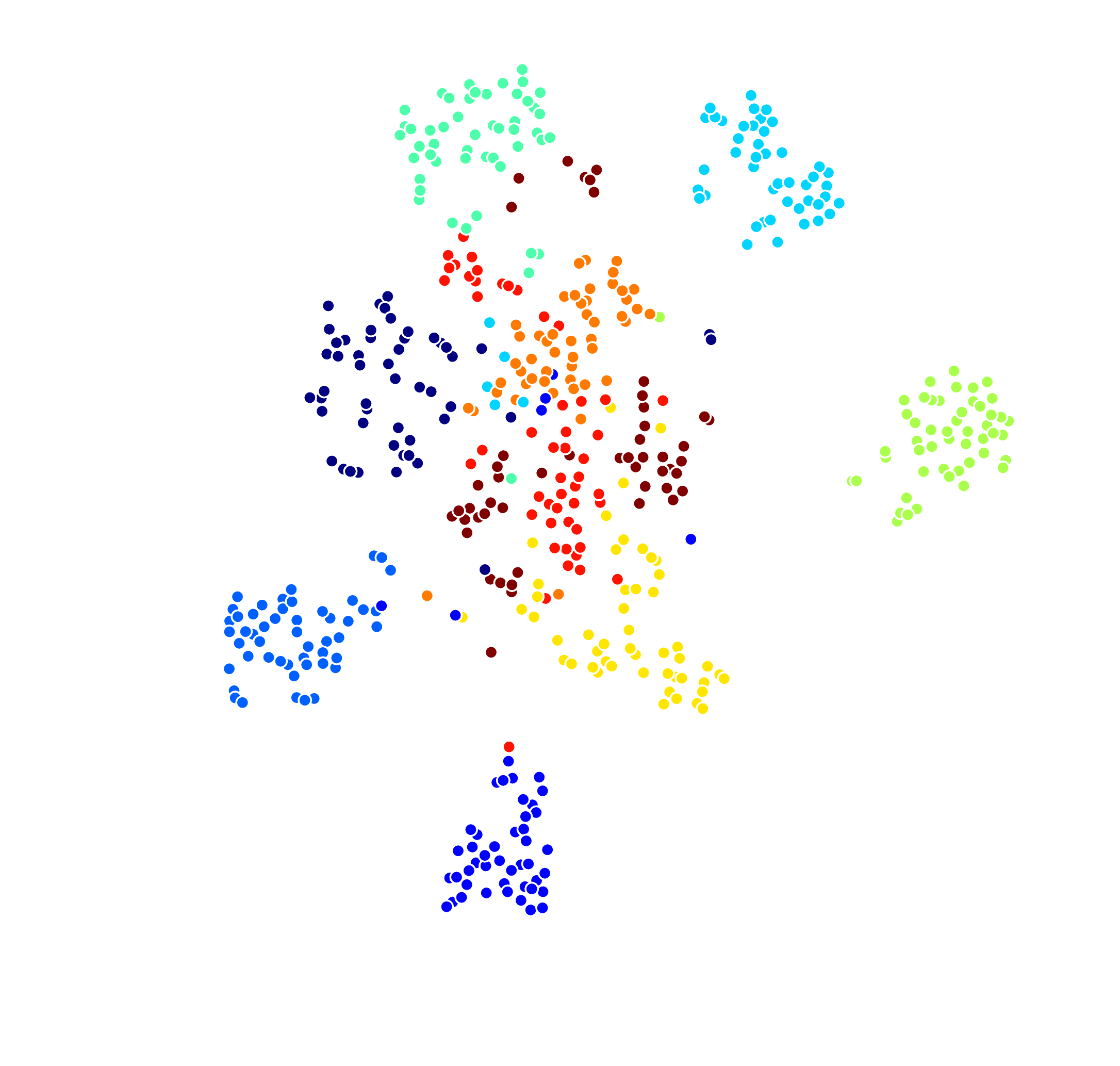}
          \caption{$\bm z=g(\bm h)$}
        \end{subfigure}\caption{\label{fig:tsne} t-SNE visualizations of hidden vectors of images from a randomly selected 10 classes in the validation set.}
    \end{minipage}
\end{figure}

Figure~\ref{fig:critic_linproj_eig} shows the eigenvalue distribution of linear projection matrix $W\in R^{2048\times2048}$ used to compute $\bm z = W\bm h$. This matrix has relatively few large eigenvalues, indicating that it is approximately low-rank.

Figure~\ref{fig:tsne} shows t-SNE~\cite{maaten2008visualizing} visualizations of $\bm h$ and $\bm z=g(\bm h)$ for randomly selected 10 classes by our best ResNet-50 (top-1 linear evaluation 69.3\%). Classes represented by $\bm h$ are better separated compared to $\bm z$.

\subsection{Semi-supervised Learning via Fine-Tuning}
\label{app:result_semi}

\paragraph{Fine-tuning Procedure} We fine-tune using the Nesterov momentum optimizer with a batch size of 4096, momentum of 0.9, and a learning rate of 0.8 (following $\mathrm{LearningRate}=0.05\times\mathrm{BatchSize}/256$) without warmup. Only random cropping (with random left-to-right flipping and resizing to 224x224) is used for preprocessing. We do not use any regularization (including weight decay). For 1\% labeled data we fine-tune for 60 epochs, and for 10\% labeled data we fine-tune for 30 epochs. For the inference, we resize the given image to 256x256, and take a single center crop of 224x224. 

Table~\ref{tab:data_efficient_sota_top1} shows the comparisons of top-1 accuracy for different methods for semi-supervised learning. Our models significantly improve state-of-the-art.

\begin{table}[h!]\small
    \centering
    \captionsetup{type=table} \begin{tabular}{@{\hspace{.2cm}}llcc@{\hspace{.2cm}}}
    \toprule
    \multirow{3}{*}{Method} & \multirow{3}{*}{Architecture} & \multicolumn{2}{c}{Label fraction} \\
    & & 1\% & 10\% \\
    & & \multicolumn{2}{c}{Top 1} \\
    \midrule
    Supervised baseline      & ResNet-50  &  25.4 & 56.4 \\ \midrule
    \multicolumn{4}{l}{\textit{Methods using label-propagation:}}         \\ 
UDA (w. RandAug)                        & ResNet-50 &  - & 68.8 \\
    FixMatch (w. RandAug)         & ResNet-50 & - & 71.5 \\
    S4L (Rot+VAT+Ent. Min.)          & ResNet-50 (4$\times$) &  - &  73.2\\ \midrule
    \multicolumn{4}{l}{\textit{Methods using self-supervised representation learning only:}}         \\ 
CPC v2 & ResNet-161($*$)            & 52.7  &  73.1 \\
    SimCLR (ours)   & ResNet-50        & 48.3
       & 	65.6   \\
    SimCLR (ours)   & ResNet-50 ($2\times$)           & 58.5  &  71.7  \\
    SimCLR (ours)    & ResNet-50 ($4\times$)           & \textbf{63.0}
       &  \textbf{74.4} \\
    \bottomrule
    \end{tabular}
    \caption{\label{tab:data_efficient_sota_top1}ImageNet top-1 accuracy of models trained with few labels. See Table~\ref{tab:data_efficient_sota} for top-5 accuracy.}
\end{table}

\subsection{Linear Evaluation}
\label{app:result_linear}

For linear evaluation, we follow similar procedure as fine-tuning (described in Appendix~\ref{app:result_semi}), except that a larger learning rate of 1.6 (following $\mathrm{LearningRate}=0.1\times\mathrm{BatchSize}/256$) and longer training of 90 epochs. Alternatively, using LARS optimizer with the pretraining hyper-parameters also yield similar results. Furthermore, we find that attaching the linear classifier on top of the base encoder (with a $\mathrm{stop\_gradient}$ on the input to linear classifier to prevent the label information from influencing the encoder) and train them simultaneously during the pretraining achieves similar performance.

\subsection{Correlation Between Linear Evaluation and Fine-Tuning}

Here we study the correlation between linear evaluation and fine-tuning under different settings of training step and network architecture.

\begin{figure}[h!]
    \centering
    \includegraphics[width=0.6\textwidth]{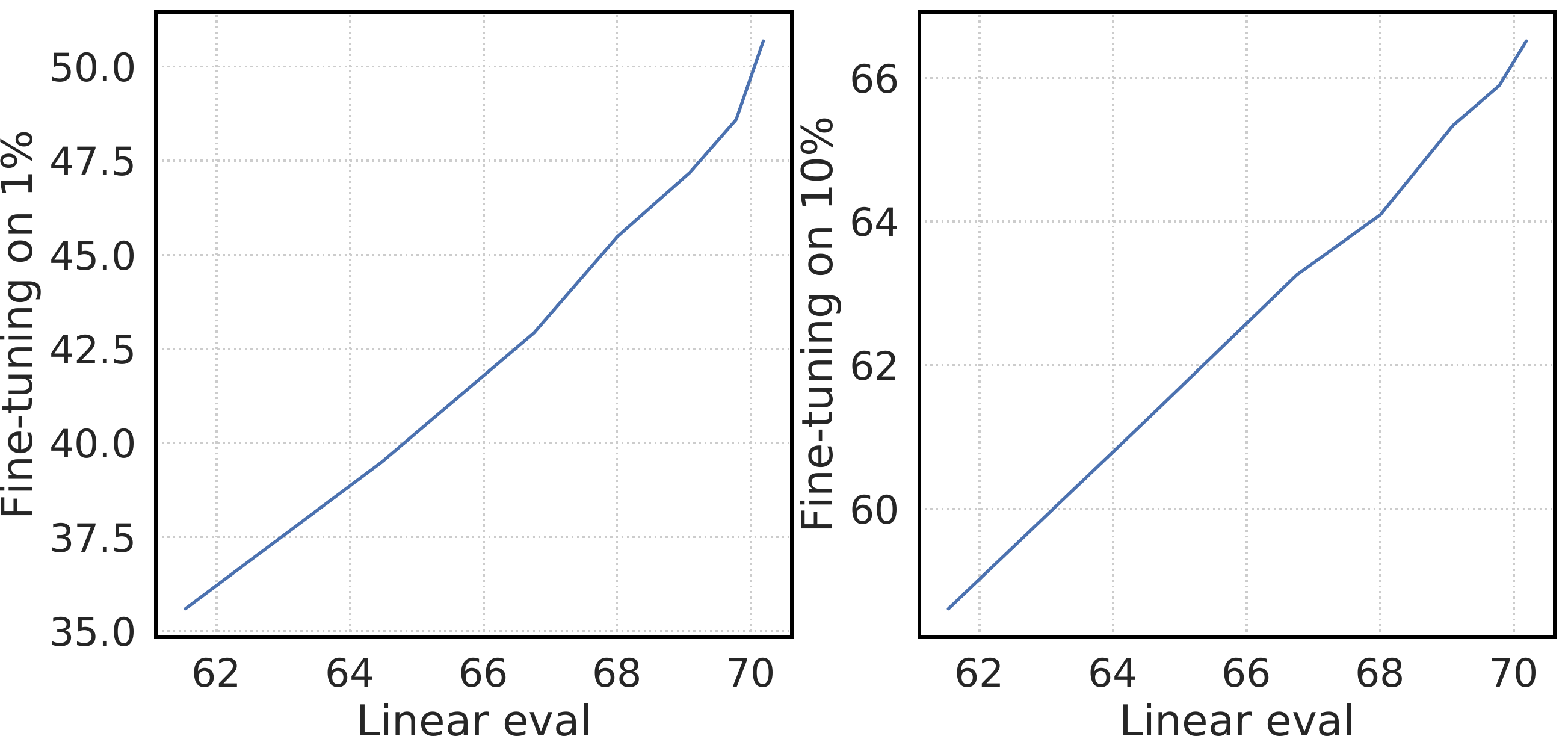}
    \caption{Top-1 accuracy of models trained in different epochs (from Figure~\ref{fig:bsstep_top1_longer}), under linear evaluation and fine-tuning.}
    \label{fig:epoch_ft_linear}
\end{figure}

Figure~\ref{fig:epoch_ft_linear} shows linear evaluation versus fine-tuning when training epochs of a ResNet-50 (using batch size of 4096) are varied from 50 to 3200 as in Figure~\ref{fig:bsstep_top1_longer}. While they are almost linearly correlated, it seems fine-tuning on a small fraction of labels benefits more from training longer.

\begin{figure}[h!]
    \centering
     \begin{subfigure}[b]{0.3\textwidth}
         \centering
         \includegraphics[width=\textwidth]{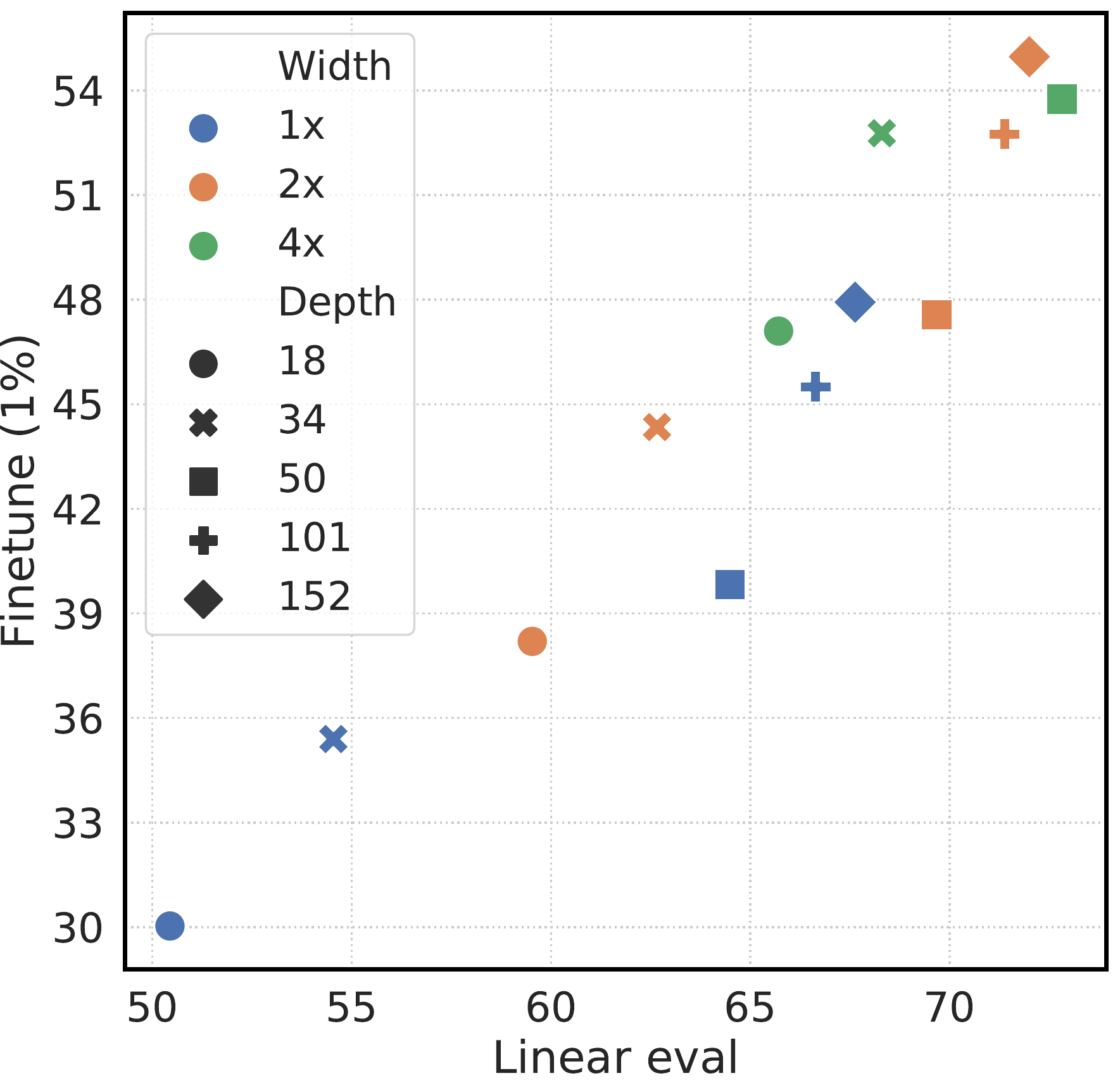}
     \end{subfigure}
     ~~~~
     \begin{subfigure}[b]{0.3\textwidth}
         \centering
         \includegraphics[width=\textwidth]{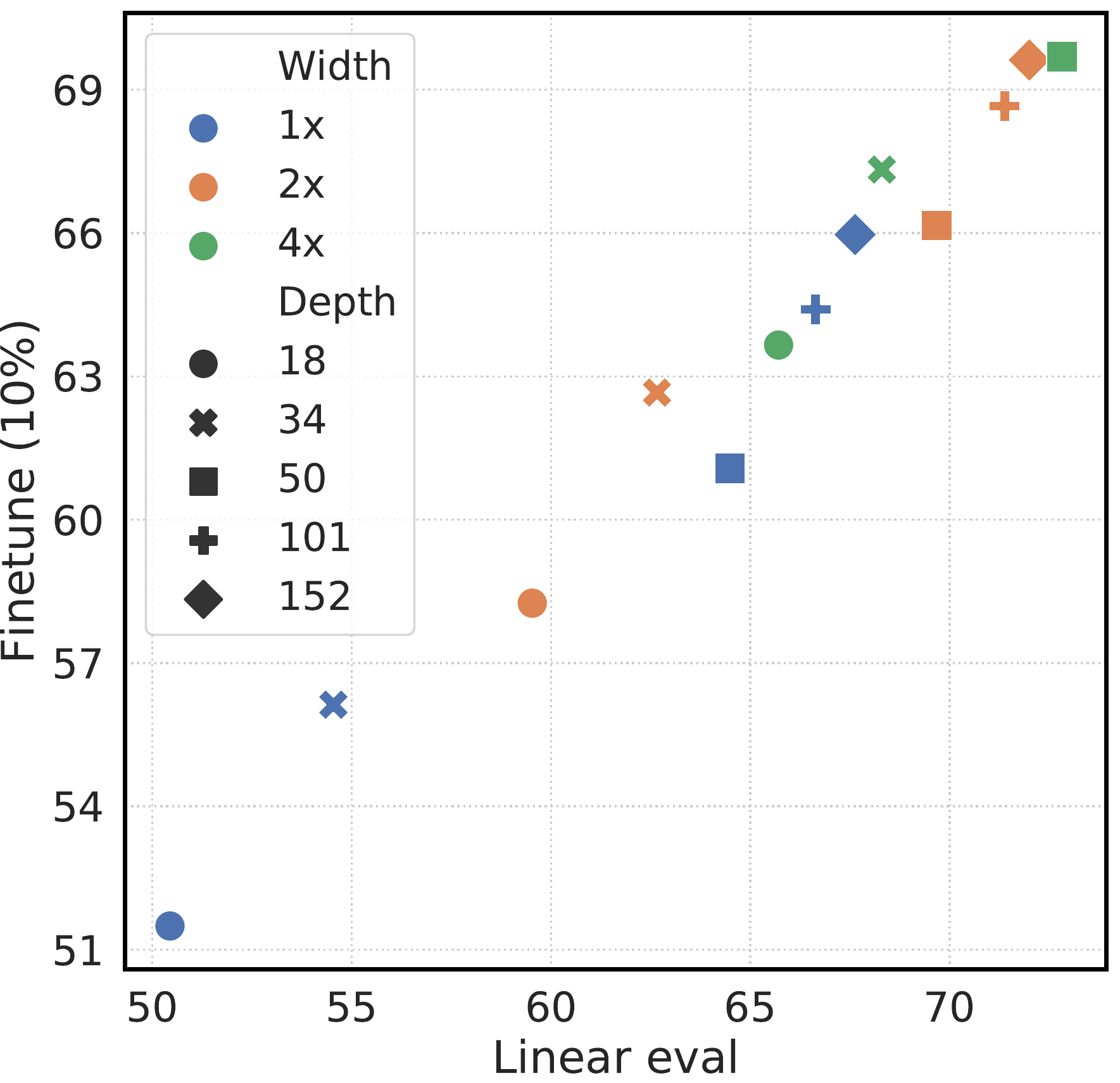}
     \end{subfigure}
    \caption{\label{fig:arch_linear_ft} Top-1 accuracy of different architectures under linear evaluation and fine-tuning.}
\end{figure}

Figure \ref{fig:arch_linear_ft} shows shows linear evaluation versus fine-tuning for different architectures of choice.

\subsection{Transfer Learning}
\label{app:transfer_learning}
We evaluated the performance of our self-supervised representation for transfer learning in two settings: linear evaluation, where a logistic regression classifier is trained to classify a new dataset based on the self-supervised representation learned on ImageNet, and fine-tuning, where we allow all weights to vary during training. In both cases, we follow the approach described by~\citet{kornblith2019better}, although our preprocessing differs slightly.

\subsubsection{Methods}

\paragraph{Datasets} We investigated transfer learning performance on the Food-101 dataset~\cite{bossard2014food}, CIFAR-10 and CIFAR-100~\cite{krizhevsky2009learning}, Birdsnap~\cite{berg2014birdsnap}, the SUN397 scene dataset~\cite{xiao2010sun}, Stanford Cars~\cite{krause2013collecting}, FGVC Aircraft~\cite{maji13fine-grained}, the PASCAL VOC 2007 classification task~\cite{everingham2010pascal}, the Describable Textures Dataset (DTD)~\cite{cimpoi2014describing}, Oxford-IIIT Pets~\cite{parkhi2012cats}, Caltech-101~\cite{fei2004learning}, and Oxford 102 Flowers~\cite{nilsback2008automated}. We follow the evaluation protocols in the papers introducing these datasets, \textit{i.e.}, we report top-1 accuracy for Food-101, CIFAR-10, CIFAR-100, Birdsnap, SUN397, Stanford Cars, and DTD; mean per-class accuracy for FGVC Aircraft, Oxford-IIIT Pets, Caltech-101, and Oxford 102 Flowers; and the 11-point mAP metric as defined in \citet{everingham2010pascal} for PASCAL VOC 2007. For DTD and SUN397, the dataset creators defined multiple train/test splits; we report results only for the first split. Caltech-101 defines no train/test split, so we randomly chose 30 images per class and test on the remainder, for fair comparison with previous work~\cite{donahue2014decaf,simonyan2014very}.

We used the validation sets specified by the dataset creators to select hyperparameters for FGVC Aircraft, PASCAL VOC 2007, DTD, and Oxford 102 Flowers. For other datasets, we held out a subset of the training set for validation while performing hyperparameter tuning. After selecting the optimal hyperparameters on the validation set, we retrained the model using the selected parameters using all training and validation images. We report accuracy on the test set.

\paragraph{Transfer Learning via a Linear Classifier} We trained an $\ell_2$-regularized multinomial logistic regression classifier on features extracted from the frozen pretrained network. We used L-BFGS to optimize the softmax cross-entropy objective and we did not apply data augmentation. As preprocessing, all images were resized to 224 pixels along the shorter side using bicubic resampling, after which we took a $224\times 224$ center crop. We selected the $\ell_2$ regularization parameter from a range of 45 logarithmically spaced values between $10^{-6}$ and $10^5$.

\paragraph{Transfer Learning via Fine-Tuning} We fine-tuned the entire network using the weights of the pretrained network as initialization. We trained for 20,000 steps at a batch size of 256 using SGD with Nesterov momentum with a momentum parameter of 0.9. We set the momentum parameter for the batch normalization statistics to $\max(1 - 10/s, 0.9)$ where $s$ is the number of steps per epoch. As data augmentation during fine-tuning, we performed only random crops with resize and flips; in contrast to pretraining, we did not perform color augmentation or blurring. At test time, we resized images to 256 pixels along the shorter side and took a $224\times 224$ center crop. (Additional accuracy improvements may be possible with further optimization of data augmentation, particularly on the CIFAR-10 and CIFAR-100 datasets.) We selected the learning rate and weight decay, with a grid of 7 logarithmically spaced learning rates between 0.0001 and 0.1 and 7 logarithmically spaced values of weight decay between $10^{-6}$ and $10^{-3}$, as well as no weight decay. We divide these values of weight decay by the learning rate. 

\paragraph{Training from Random Initialization} We trained the network from random initialization using the same procedure as for fine-tuning, but for longer, and with an altered hyperparameter grid. We chose hyperparameters from a grid of 7 logarithmically spaced learning rates between 0.001 and 1.0 and 8 logarithmically spaced values of weight decay between $10^{-5}$ and $10^{-1.5}$. 
Importantly, our random initialization baselines are trained for 40,000 steps, which is sufficiently long to achieve near-maximal accuracy, as demonstrated in Figure 8 of~\citet{kornblith2019better}.

On Birdsnap, there are no statistically significant differences among methods, and on Food-101, Stanford Cars, and FGVC Aircraft datasets, fine-tuning provides only a small advantage over training from random initialization. However, on the remaining 8 datasets, pretraining has clear advantages.

\paragraph{Supervised Baselines} We compare against architecturally identical ResNet models trained on ImageNet with standard cross-entropy loss. These models are trained with the same data augmentation as our self-supervised models (crops, strong color augmentation, and blur) and are also trained for 1000 epochs. We found that, although stronger data augmentation and longer training time do not benefit accuracy on ImageNet, these models performed significantly better than a supervised baseline trained for 90 epochs and ordinary data augmentation for linear evaluation on a subset of transfer datasets. The supervised ResNet-50 baseline achieves 76.3\% top-1 accuracy on ImageNet, vs. 69.3\% for the self-supervised counterpart, while the ResNet-50 ($4 \times$) baseline achieves 78.3\%, vs. 76.5\% for the self-supervised model.

\paragraph{Statistical Significance Testing} We test for the significance of differences between model with a permutation test. Given predictions of two models, we generate 100,000 samples from the null distribution by randomly exchanging predictions for each example and computing the difference in accuracy after performing this randomization. We then compute the percentage of samples from the null distribution that are more extreme than the observed difference in predictions. For top-1 accuracy, this procedure yields the same result as the exact McNemar test. The assumption of exchangeability under the null hypothesis is also valid for mean per-class accuracy, but not when computing average precision curves. Thus, we perform significance testing for a difference in accuracy on VOC 2007 rather than a difference in mAP. A caveat of this procedure is that it does not consider run-to-run variability when training the models, only variability arising from using a finite sample of images for evaluation.

\subsubsection{Results with Standard ResNet}

\begin{table*}[h]
\footnotesize
\centering
\setlength{\tabcolsep}{3pt}
\begin{tabular}{lcccccccccccc}
\toprule
{} &  Food &  CIFAR10 &  CIFAR100 &  Birdsnap &  SUN397 &  Cars &  Aircraft &  VOC2007 &  DTD &  Pets &  Caltech-101 &  Flowers \\
\midrule
\multicolumn{5}{l}{\textit{Linear evaluation:}}\\
SimCLR (ours)    &     68.4 &     90.6 &      71.6 &      37.4 &    58.8 &           50.3 &           50.3 &     80.5 &        \textbf{74.5} &         83.6 &               90.3 &               91.2 \\
Supervised &     \textbf{72.3} &     \textbf{93.6} &      \textbf{78.3} &      \textbf{53.7} &    \textbf{61.9} &           \textbf{66.7} &           \textbf{61.0} &     \textbf{82.8} &        \textbf{74.9} &         \textbf{91.5} &               \textbf{94.5} &               \textbf{94.7} \\
\midrule
\multicolumn{5}{l}{\textit{Fine-tuned:}}\\
SimCLR (ours)    &     \textbf{88.2} &     \textbf{97.7} &      \textbf{85.9} &      \textbf{75.9} &    63.5 &           91.3 &           \textbf{88.1} &     84.1 &        \textbf{73.2} &         89.2 &               92.1 &               97.0 \\
    Supervised &     \textbf{88.3} &     \textbf{97.5} &      \textbf{86.4} &      \textbf{75.8} &    \textbf{64.3} &           \textbf{92.1} &           86.0 &     \textbf{85.0} &        \textbf{74.6} &         \textbf{92.1} &               \textbf{93.3} &               \textbf{97.6}\\
Random init
&     86.9 &     95.9 &      80.2 &      \textbf{76.1} &    53.6 &           91.4 &           85.9 &     67.3 &        64.8 &         81.5 &               72.6 &               92.0 \\
\bottomrule
\end{tabular}
\caption{Comparison of transfer learning performance of our self-supervised approach with supervised baselines across 12 natural image datasets, using ImageNet-pretrained ResNet models. See also Figure~\ref{tab:transfer_learning_resnet_4x} for results with the ResNet ($4 \times$) architecture.}
\label{tab:transfer_learning_resnet}
\end{table*}

The ResNet-50 ($4\times$) results shown in Table~\ref{tab:transfer_learning_resnet_4x} of the text show no clear advantage to the supervised or self-supervised models. With the narrower ResNet-50 architecture, however, supervised learning maintains a clear advantage over self-supervised learning. The supervised ResNet-50 model outperforms the self-supervised model on all datasets with linear evaluation, and most (10 of 12) datasets with fine-tuning. The weaker performance of the ResNet model compared to the ResNet ($4\times$) model may relate to the accuracy gap between the supervised and self-supervised models on ImageNet. The self-supervised ResNet gets 69.3\% top-1 accuracy, 6.8\% worse than the supervised model in absolute terms, whereas the self-supervised ResNet $(4\times)$ model gets 76.5\%, which is only 1.8\% worse than the supervised model.

\subsection{CIFAR-10}
\label{app:cifar}
While we focus on using ImageNet as the main dataset for pretraining our unsupervised model, our method also works with other datasets. We demonstrate it by testing on CIFAR-10 as follows.

\paragraph{Setup} As our goal is not to optimize CIFAR-10 performance, but rather to provide further confirmation of our observations on ImageNet, we use the same architecture (ResNet-50) for CIFAR-10 experiments. Because CIFAR-10 images are much smaller than ImageNet images, we replace the first 7x7 Conv of stride 2 with 3x3 Conv of stride 1, and also remove the first max pooling operation. For data augmentation, we use the same Inception crop (flip and resize to 32x32) as ImageNet,\footnote{It is worth noting that, although CIFAR-10 images are much smaller than ImageNet images and image size does not differ among examples, cropping with resizing is still a very effective augmentation for contrastive learning.} and color distortion (strength=0.5), leaving out Gaussian blur. We pretrain with learning rate in $\{0.5, 1.0, 1.5\}$, temperature in $\{0.1, 0.5, 1.0\}$, and batch size in $\{256, 512, 1024, 2048, 4096\}$. The rest of the settings (including optimizer, weight decay, etc.) are the same as our ImageNet training.

Our best model trained with batch size 1024 can achieve a linear evaluation accuracy of 94.0\%, compared to 95.1\% from the supervised baseline using the same architecture and batch size. The best self-supervised model that reports linear evaluation result on CIFAR-10 is AMDIM~\cite{bachman2019learning}, which achieves 91.2\% with a model $25\times$ larger than ours. We note that our model can be improved by incorporating extra data augmentations as well as using a more suitable base network.

\begin{figure}[h!]
\centering
\begin{minipage}[t]{0.5\textwidth}
\vspace{0pt}
\includegraphics[width=\linewidth]{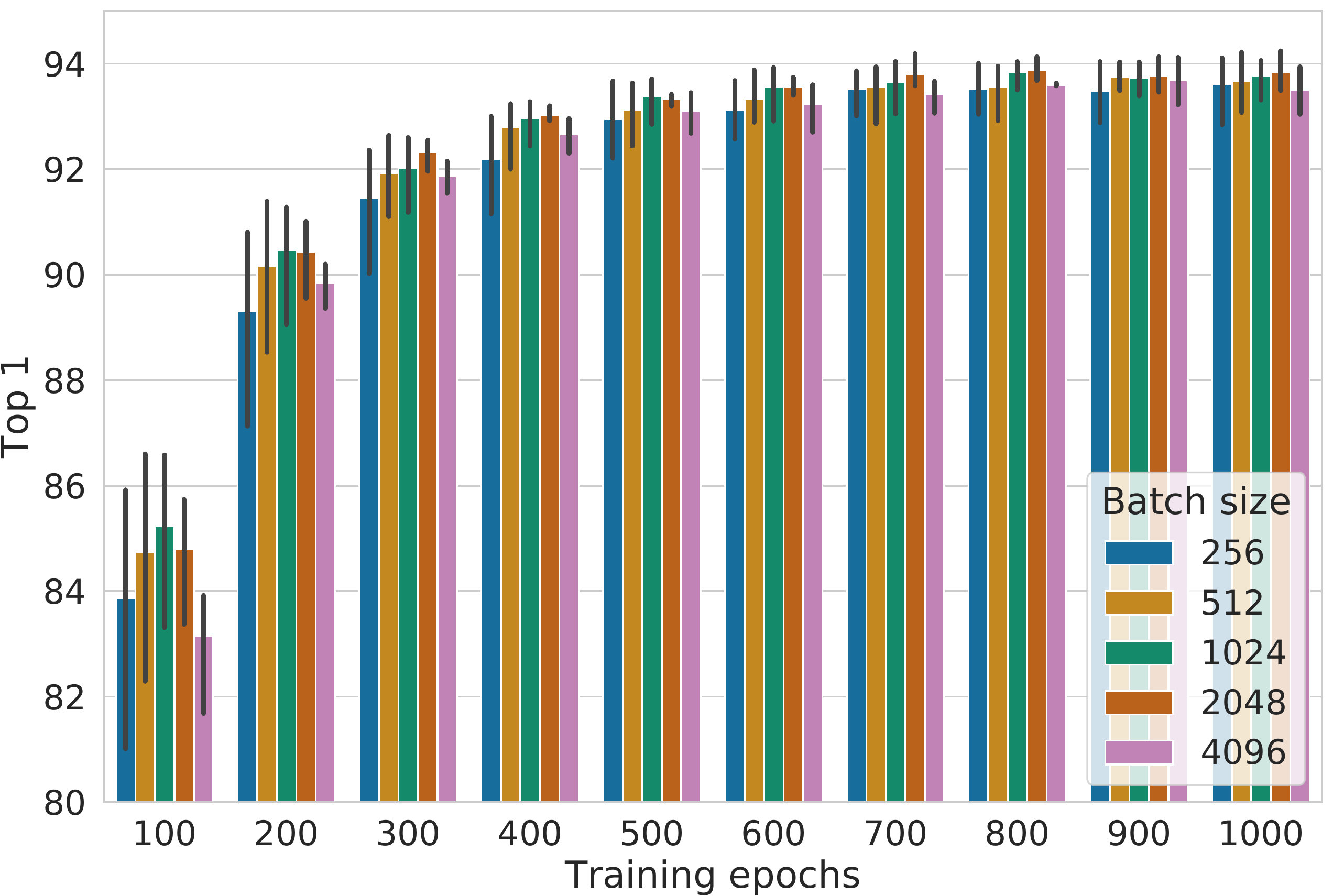}
\end{minipage}\hspace{0.02\textwidth}
\begin{minipage}[t]{0.38\textwidth}
\vspace{0pt}
\caption{\label{fig:bsstep_cifar}Linear evaluation of ResNet-50 (with adjusted stem) trained with different batch size and epochs on CIFAR-10 dataset. Each bar is averaged over 3 runs with different learning rates (0.5, 1.0, 1.5) and temperature $\tau=0.5$. Error bar denotes standard deviation.}
\end{minipage}
\end{figure}

\paragraph{Performance under different batch sizes and training steps} Figure \ref{fig:bsstep_cifar} shows the linear evaluation performance under different batch sizes and training steps. The results are consistent with our observations on ImageNet, although the largest batch size of 4096 seems to cause a small degradation in performance on CIFAR-10.

\begin{figure}[h!]
    \centering
    \begin{subfigure}{.3\textwidth}
      \centering
      \includegraphics[width=\linewidth]{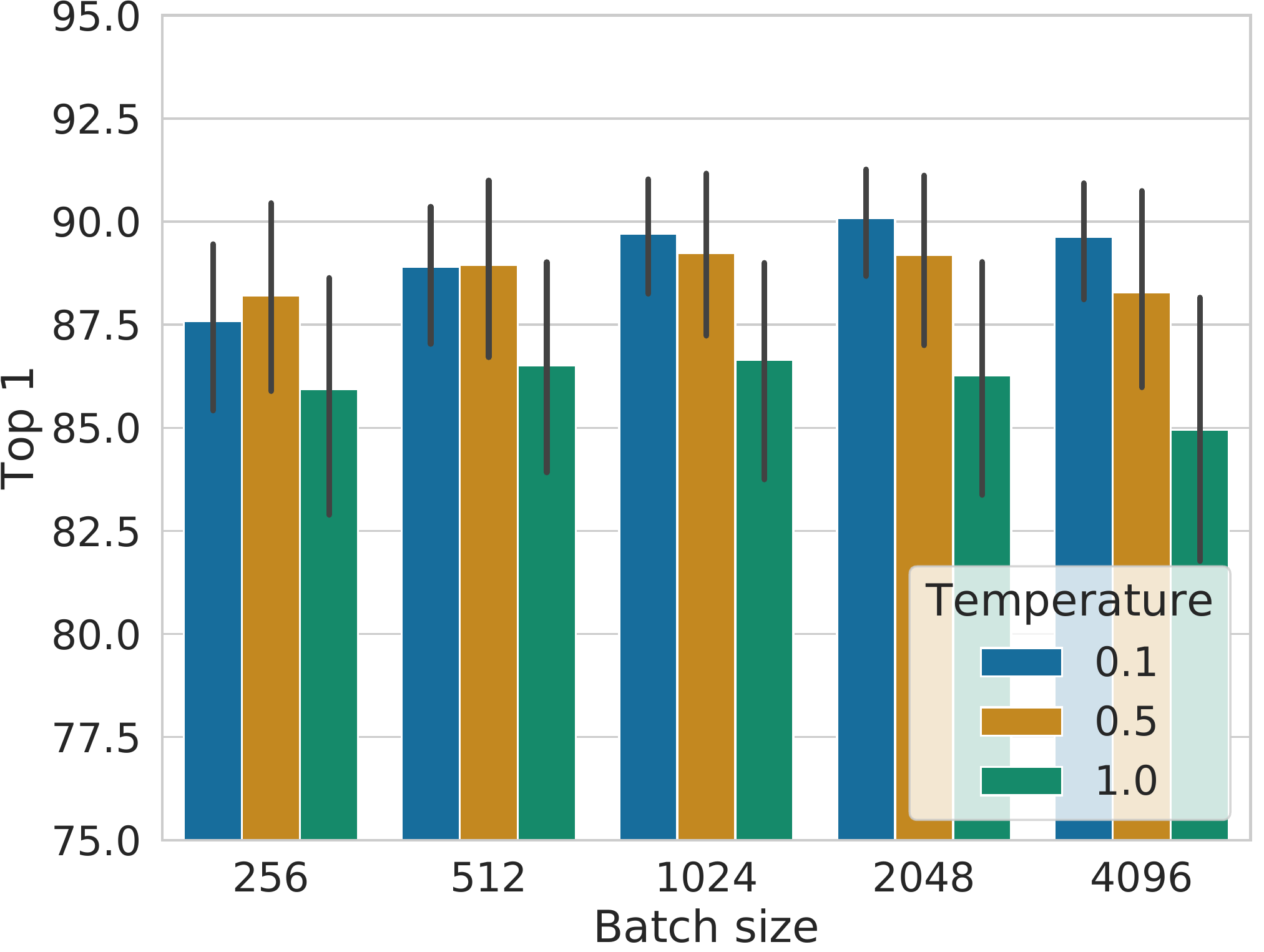}
      \caption{Training epochs $\le$ 300}
    \end{subfigure}~~~~
    \begin{subfigure}{.3\textwidth}
      \centering
      \includegraphics[width=\linewidth]{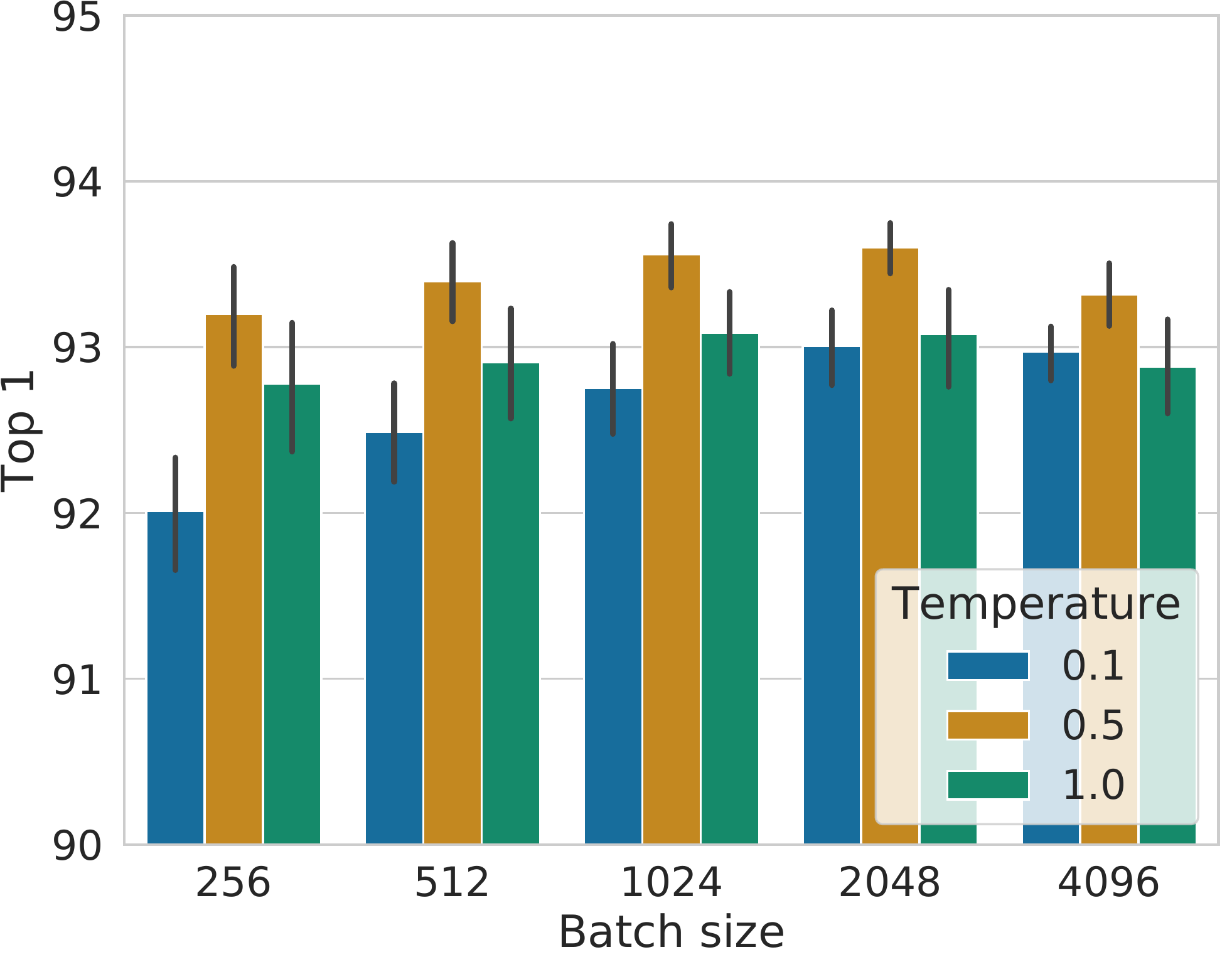}
      \caption{Training epochs $>$ 300}
    \end{subfigure}\caption{\label{fig:bstau_cifar}Linear evaluation of the model (ResNet-50) trained with three temperatures on different batch sizes on CIFAR-10. Each bar is averaged over multiple runs with different learning rates and total train epochs. Error bar denotes standard deviation.}
\end{figure}

\paragraph{Optimal temperature under different batch sizes} Figure~\ref{fig:bstau_cifar} shows the linear evaluation of model trained with three different temperatures under various batch sizes. We find that when training to convergence (e.g. training epochs > 300), the optimal temperature in $\{0.1, 0.5, 1.0\}$ is 0.5 and seems consistent regardless of the batch sizes. However, the performance with $\tau=0.1$ improves as batch size increases, which may suggest a small shift of optimal temperature towards 0.1.

\subsection{Tuning For Other Loss Functions}
\label{app:loss_tuning}

The learning rate that works best for NT-Xent loss may not be a good learning rate for other loss functions. To ensure a fair comparison, we also tune hyperparameters for both margin loss and logistic loss. Specifically, we tune learning rate in $\{0.01, 0.1, 0.3, 0.5, 1.0\}$ for both loss functions. We further tune the margin in $\{0, 0.4, 0.8, 1.6\}$ for margin loss, the temperature in $\{0.1, 0.2, 0.5, 1.0\}$ for logistic loss. For simplicity, we only consider the negatives from one augmentation view (instead of both sides), which slightly impairs performance but ensures fair comparison.

\section{Further Comparison to Related Methods}
\label{app:related}

As we have noted in the main text, most individual components of SimCLR have appeared in previous work, and the improved performance is a result of a combination of these design choices. Table \ref{tab:comp_design} provides a high-level comparison of the design choices of our method with those of previous methods.
Compared with previous work, our design choices are generally simpler.

\begin{table}[h!]
\small
\centering
\begin{tabular}{lllllll}
\toprule
Model & Data Augmentation & Base Encoder & Projection Head & Loss & Batch Size & Train Epochs  \\\midrule 
CPC v2 & Custom & ResNet-161 (modified) & PixelCNN & Xent & 512$^\#$ & $\sim$200 \\
AMDIM & Fast AutoAug. & Custom ResNet  & Non-linear MLP & Xent w/ clip,reg & 1008$^\#$ & 150 \\
CMC & Fast AutoAug.  & ResNet-50 ($2\times$, L+ab)  & Linear layer & Xent w/ $\ell_2,\tau$ & 156$^*$ & 280 \\
MoCo & Crop+color  & ResNet-50 (4$\times$)  & Linear layer & Xent w/ $\ell_2,\tau$ & 256$^*$ & 200 \\
PIRL &  Crop+color & ResNet-50 (2$\times$) & Linear layer & Xent w/ $\ell_2,\tau$ & 1024$^*$ & 800 \\
SimCLR & Crop+color+blur  & ResNet-50 ($4\times$)  & Non-linear MLP  & Xent w/ $\ell_2,\tau$ & 4096 & 1000 \\
\bottomrule
\end{tabular}
\caption{\label{tab:comp_design} A high-level comparison of design choices and training setup (for best result on ImageNet) for each method. Note that descriptions provided here are general; even when they match for two methods, formulations and implementations may differ (e.g. for color augmentation). Refer to the original papers for more details. $^\#$Examples are split into multiple patches, which enlarges the effective batch size. $^*$A memory bank is employed.}
\end{table}

In below, we provide an in-depth comparison of our method to the recently proposed contrastive representation learning methods:
\begin{itemize}
    \item DIM/AMDIM~\cite{hjelm2018learning,bachman2019learning} achieve global-to-local/local-to-neighbor prediction by predicting the middle layer of ConvNet. The ConvNet is a ResNet that has bewen modified to place significant constraints on the receptive fields of the network (e.g. replacing many 3x3 Convs with 1x1 Convs). In our framework, we decouple the prediction task and encoder architecture, by random cropping (with resizing) and using the final representations of two augmented views for prediction, so we can use standard and more powerful ResNets. Our NT-Xent loss function leverages normalization and temperature to restrict the range of similarity scores, whereas they use a tanh function with regularization. We use a simpler data augmentation policy, while they use FastAutoAugment for their best result.
    \item CPC v1 and v2~\cite{oord2018representation,henaff2019data} define the context prediction task using a deterministic strategy to split examples into patches, and a context aggregation network (a PixelCNN) to aggregate these patches. The base encoder network sees only patches, which are considerably smaller than the original image. We decouple the prediction task and the encoder architecture, so we do not require a context aggregation network, and our encoder can look at the images of wider spectrum of resolutions. In addition, we use the NT-Xent loss function, which leverages normalization and temperature, whereas they use an unnormalized cross-entropy-based objective. We use simpler data augmentation.
    \item InstDisc, MoCo, PIRL~\cite{wu2018unsupervised,he2019momentum,misra2019self} generalize the Exemplar approach originally proposed by~\citet{dosovitskiy2014discriminative} and leverage an explicit memory bank. We do not use a memory bank; we find that, with a larger batch size, in-batch negative example sampling suffices. We also utilize a nonlinear projection head, and use the representation before the projection head. Although we use similar types of augmentations (e.g., random crop and color distortion), we expect specific parameters may be different.
    \item CMC~\cite{tian2019contrastive} uses a separated network for each view, while we simply use a single network shared for all randomly augmented views. The data augmentation, projection head and loss function are also different. We use larger batch size instead of a memory bank.
    \item Whereas \citet{ye2019unsupervised} maximize similarity between augmented and unaugmented copies of the same image, we apply data augmentation symmetrically to both branches of our framework (Figure~\ref{fig:framework}). We also apply a nonlinear projection on the output of base feature network, and use the representation before projection network, whereas \citet{ye2019unsupervised} use the linearly projected final hidden vector as the representation. When training with large batch sizes using multiple accelerators, we use global BN to avoid shortcuts that can greatly decrease representation quality.
\end{itemize}

\end{document}